\documentclass[sn-mathphys]{sn-jnl}% Math and Physical Sciences Reference Style
\usepackage{xcolor}
\usepackage{subfig}
\usepackage{pythonhighlight}
\usepackage{listings}
\usepackage{mathtools}
\usepackage{commath}

\jyear{2022}%

\definecolor{amaranth}{rgb}{0.9, 0.17, 0.31}
\definecolor{antiquefuchsia}{rgb}{0.57, 0.36, 0.51}
\definecolor{amethyst}{rgb}{0.6, 0.4, 0.8}
\definecolor{deepblue}{rgb}{0.07, 0.21, 0.58}

\theoremstyle{thmstyleone}%
\newcommand{\vect}[1]{\boldsymbol{#1}}

\theoremstyle{thmstyletwo}%

\theoremstyle{thmstylethree}%

\newcommand{\obs}{o}
\newcommand{\svar}{s}
\newcommand{\params}{w}
\newcommand{\pred}{\hat{v}}
\newcommand{\shat}{\hat{s}}

\begin{document}

\title[GVFs in the Real World: Making Predictions Online for Water Treatment
]{GVFs in the Real World: Making Predictions Online for Water Treatment}

%%=============================================================%%
%% Prefix	-> \pfx{Dr}
%% GivenName	-> \fnm{Joergen W.}
%% Particle	-> \spfx{van der} -> surname prefix
%% FamilyName	-> \sur{Ploeg}
%% Suffix	-> \sfx{IV}
%% NatureName	-> \tanm{Poet Laureate} -> Title after name
%% Degrees	-> \dgr{MSc, PhD}
%% \author*[1,2]{\pfx{Dr} \fnm{Joergen W.} \spfx{van der} \sur{Ploeg} \sfx{IV} \tanm{Poet Laureate} 
%%                 \dgr{MSc, PhD}}\email{iauthor@gmail.com}
%%=============================================================%%

\author*[1]{\fnm{Muhammad Kamran} \sur{Janjua}}\email{mjanjua@ualberta.ca}

\author[1]{\fnm{Haseeb} \sur{Shah}}

\author[1,2]{\fnm{Martha} \sur{White}}

\author[1]{\fnm{Erfan} \sur{Miahi}}

\author[1,2]{\fnm{Marlos} \sur{C. Machado}}

\author*[1,2]{\fnm{Adam} \sur{White}}\email{amw8@ualberta.ca}

\affil[1]{\orgdiv{Department of Computing Science},  Alberta Machine Intelligence Institute (Amii), \orgname{University of Alberta}, \orgaddress{\city{Edmonton},\country{Canada}}}
\affil[2]{\orgdiv{Canada CIFAR AI Chair}}

%\author*[1,2]{\fnm{First} \sur{Author}}\email{iauthor@gmail.com}
%
%\author[2,3]{\fnm{Second} \sur{Author}}\email{iiauthor@gmail.com}
%\equalcont{These authors contributed equally to this work.}
%
%\author[1,2]{\fnm{Third} \sur{Author}}\email{iiiauthor@gmail.com}
%\equalcont{These authors contributed equally to this work.}
%
%\affil*[1]{\orgdiv{Department}, \orgname{Organization}, \orgaddress{\street{Street}, \city{City}, \postcode{100190}, \state{State}, \country{Country}}}
%
%\affil[2]{\orgdiv{Department}, \orgname{Organization}, \orgaddress{\street{Street}, \city{City}, \postcode{10587}, \state{State}, \country{Country}}}
%
%\affil[3]{\orgdiv{Department}, \orgname{Organization}, \orgaddress{\street{Street}, \city{City}, \postcode{610101}, \state{State}, \country{Country}}}

%\abstract{Time series prediction is a typical approach to making predictions on temporal data. In this work, we highlight how RL prediction approaches can provide a useful alternative to the standard time series questions and algorithms. In particular, we highlight the utility of (a) focusing on creating agent state to improve prediction accuracy, (b) using cumulative targets as opposed to n-horizon predictions, and (c) updating predictions online, facilitated by the bootstrap updates in RL algorithms. We provide a case study using data generated from a real drinking water treatment system. The dataset is a high-dimensional, noisy time series dataset. We release this dataset, as well as associated code for learning and evaluating predictions.}
\abstract{In this paper we investigate the use of reinforcement-learning based prediction approaches for a real drinking-water treatment plant. Developing such a prediction system is a critical step on the path to optimizing and automating water treatment. Before that, there are many questions to answer about the predictability of the data, suitable neural network architectures, how to overcome partial observability and more. We first describe this dataset, and highlight challenges with seasonality, nonstationarity, partial observability, and heterogeneity across sensors and operation modes of the plant. We then describe General Value Function (GVF) predictions---discounted cumulative sums of observations--and highlight why they might be preferable to classical n-step predictions common in time series prediction. We discuss how to use offline data to appropriately pre-train our temporal difference learning (TD) agents that learn these GVF predictions, including how to select hyperparameters for online fine-tuning in deployment. We find that the TD-prediction agent obtains an overall lower normalized mean-squared error than the n-step prediction agent. Finally, we show the importance of learning in deployment, by comparing a TD agent trained purely offline with no online updating to a TD agent that learns online. This final result is one of the first to motivate the importance of adapting predictions in real-time, for non-stationary high-volume systems in the real world.} 

\maketitle

\section{Introduction}\label{sec:intro}
Learning in deployment is critical for partially observable decision-making tasks~\cite{sutton2007role}. If the evolution of state transitions is driven by both the agent’s actions and state variables that the agent cannot observe, then the process will appear non-stationary to the agent. For example, an agent controlling chemical dosing in a water treatment plant may correctly learn the relationship between increasing chemicals to reduce turbidity in the water. However, inclement weather events can also impact water turbidity causing the agent’s prediction of future turbidity—and thus choices of chemical dosing—to be suboptimal. One approach to mitigate this problem is to allow the agent to continually update its predictions and decision-making policies online in deployment.

Effective multi-step prediction forms the basis for decision-making in almost any reinforcement learning system. Classical value-based methods, such as Q-learning~\cite{watkins92q}, construct a prediction of future discounted rewards in order to decide on what actions to take. Policy gradient methods such as PPO~\cite{schulman2017proximal} and SAC~\cite{haarnoja2018soft} typically define the agent’s policy through an estimate of the value function. Thus in applications, a reasonable first step is to build a prediction learning system that can predict reward and sensor values far into the future. This is not only an important step to assess the feasibility of adaptive control but is also a useful first step because the tasks of feature engineering, network architecture design, optimization, and tuning of various hyperparameters will be shared and beneficial to both a prediction learning system and a full reinforcement learning control system.  

There has been growing interest in moving RL techniques out of video games and into the real world. In many applications, such as chip design \cite{mirhoseini2020chip}, matrix multiplication \cite{fawzi2022discovering}, and even video compression \cite{mandhane2022muzero}, the problem setting of interest is simulation. Another approach is to design and train an agent in simulation and then deploy a fixed controller, sometimes even in the real world. This approach has been used for example in navigating stratospheric balloons \cite{bellemare2020autonomous}, controlling plasma configurations inside a fusion reactor \cite{degrave2022magnetic}, and robotic curling \cite{won2020adaptive}.

In this paper, we study the application of machine learning techniques, specifically prediction methods from reinforcement learning, on a real drinking-water treatment plant. In our setting, we do not have access to a high-fidelity simulator of the plant, nor the resources to commission one. This work explores the feasibility of adaptive learning systems in the real world. Instead of relying on access to a simulator our approach extensively leverages offline data for iterating design choices or pre-training the agent. 

Closer to our work, recent work on automating HVAC control used an approach where the agent is first tuned on off-line data and then a learning controller is deployed that is updated once a day \cite{luo2022controlling}. In this work the authors explicitly avoided offline training on operator data, citing the well-known issues of insufficient action coverage. Nevertheless, {\em batch} or {\em off-line RL} methods~\cite{ernst2005tree,riedmiller2005neural,lange2012batch,levine2020offline} have been successfully used in settings where a fixed policy or value function is extracted from a data-set, with several practical applications~\cite{pietquin2011sample,shortreed2011informing,swaminathan2017off,levine2018learning}.   

\begin{figure}[!t]
	\centering
	\includegraphics[scale=0.3]{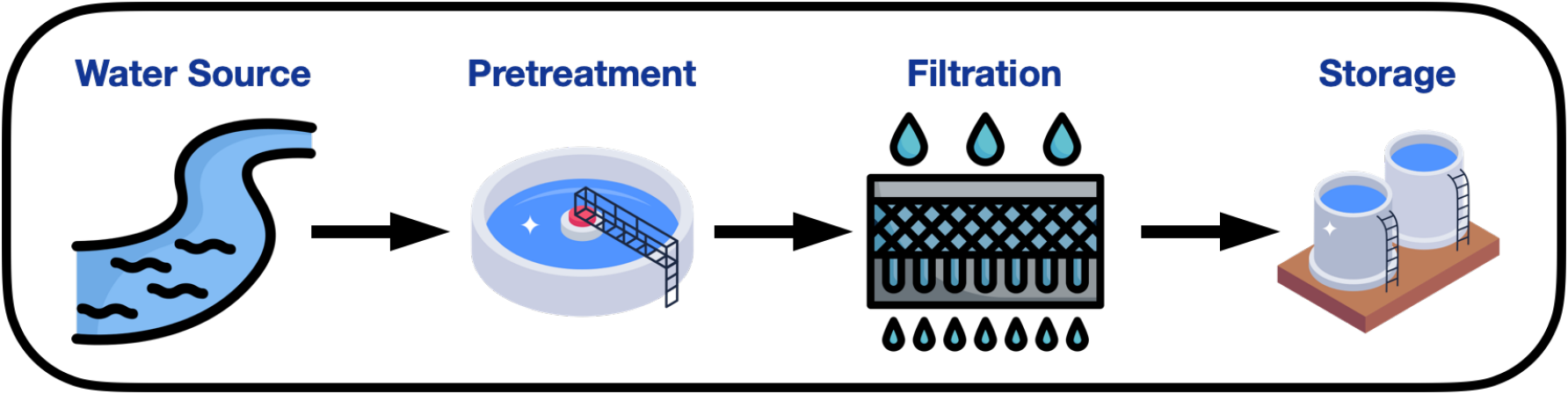}
	\caption{An illustration of the drinking-water treatment plant. The entire plant is divided into two main stages: pretreatment and filtration. The pretreatment stage is concerned with adding chemicals to the raw water, followed by the filtration stage where the water is pumped through filters for further purification.}
	\label{fig:fwt101}
\end{figure}
Drinking-water treatment is basically a two-stage process, as depicted in Figure~\ref{fig:fwt101}. First, water is pumped into a large mixing tank where chemicals are added to cause dissolved solids to clump together. The next step is to pull the pretreated water through a filter membrane where only clean water molecules can pass through the filter membrane and the solids and other continents remain. Periodically, the primary filter is cleaned by simply running the process backwards blasting the filter membrane clean: a process called backwashing. In Canada, the operation of a water-treatment plant can represent up to 30\% of a town's municipal budget \cite{copeland2014energy}. 

Drinking-water treatment is uniquely challenging compared to other applications due to two key characteristics. The data produced by a water-treatment plant, like many real-world systems, is high-dimensional, noisy, partially observable, and often incomplete, making online, continual prediction extremely challenging. In water treatment, the plant can operate in different modes, such as production and backwashing. The mode has a profound impact on data produced by the system and even changes the range of valid sensor readings. Second, the different components of the plant operate at different timescales and decisions have delayed consequences. For example, the chemical dosing rate is typically not changed more often than once a day, backwashing happens multiple times a day, and the pretreatment-tank mixing-rate can be adjusted continuously. Each one of these choices can result in changes in sensor readings over minutes—chemical dosing changes the water pressure on the filter within 30 minutes—to months—too much chemical dosing can degrade filter efficiency over the long run.   

In this paper, we investigate multi-variate, multi-step prediction in deployment on a real system. We provide a detailed case study on water treatment, first demonstrating the inherent nonstationarity of the problem and the benefits of learning continuously in deployment. We show that using a simple trace-based memory to overcome partial observability, we can learn accurate multi-step predictions, called general value functions (GVFs) \cite{sutton2011horde,modayil2014multi}, using temporal difference (TD) learning. Because GVFs can be learned with standard reinforcement learning algorithms like TD, they can easily be updated online, on every step. We show that updating online can significantly improve performance over only training from an offline log of data. The online prediction agent also benefits from this offline data, to pre-train the predictions and to set the hyperparameters for updating online in deployment. Our approach allows us to have a fully specified online prediction agent---with hyperparameters automatically selected using a simple modification on the standard validation procedure---that continues to adapt and improve in deployment. 

Finally, we also contrast these GVF multi-step predictions to the more classical predictions considered in time series prediction: n-step predictions. The primary goal of this comparison is to provide intuition: n-step predictions are a more common and widely understood approach to multi-step prediction, as compared to GVFs. Our goal is to introduce GVF predictions to a wider audience, and hopefully motivate this additional modelling tool. Beyond this, we highlight that GVFs can have benefits over n-step predictions. The target for a GVF is typically smoother because it is an exponential weighting of future observations, rather than an observation at exactly n steps in the future. Consequently, we also expect this target to be lower variance and potentially simpler to learn. We find that GVF predictions have higher accuracy than the n-step predictions on our data, controlling for the same state encoding and network size, in terms of the normalized mean-squared error. Taken together, our work provides several practical insights on designing neural-network learning systems capable of learning in deployment.%, supported by real data generated by a real water-treatment plant.

\section{The data of water treatment}\label{sec:data_description}
Like any industrial control process, a water treatment plant has the potential to generate an immense amount of data. Our system is instrumented with a large number of sensors reporting both (1) water chemistry throughout the treatment pipeline, and (2) properties of the mechanical components of the plant. Taken together these sensor readings form a long and wide time series with several interesting properties that make long-term prediction difficult. In this section we highlight these properties with examples from a real plant, explaining how each makes long-term prediction challenging.

\subsection{Wide, long, and fast data}
Our system reports 480 distinct sensor values at a rate of one reading per second producing a large time series. One year of data consists of over 31 million observations of the plant and over 15 million individual sensor readings. In contrast, the recent M5 time-series forecasting competition used a dataset with 42,840-dimensional observations and 1969 time-steps; over 84 million samples~\cite{makridakis2022m5}. Using multiple years of water treatment data puts us on the same scale as state-of-the-art forecasting grande challenge problems. We summarize some of the sensors in Table \ref{tab:summary}, and provide more detail in \ref{tab:sp_table}.

\begin{table}[t]
\centering
\caption{A brief summary of different measurements each of the sensor type is responsible for working out.}
\begin{tabular}{|l|l|l|}
\hline
\textbf{Sensor Type} & \textbf{Measures} \\ \hline
Pressure & Pressure on the membrane. \\ \hline
Flowmeter & Flow rate of the fluid. \\ \hline
pH & Acidity and alkalinity of the solution. \\ \hline
Temperature & Temperature of the water. \\ \hline
Turbidity  & Turbidity of the water. \\ \hline
Total Organic Carbon (TOC) & Organic carbon in the water. \\ \hline
Conductivity  & Ability to pass an electric current. \\ \hline
\end{tabular}
\label{tab:summary}
\end{table}

\begin{figure}[t]
	\centering
	\includegraphics[width=\textwidth]{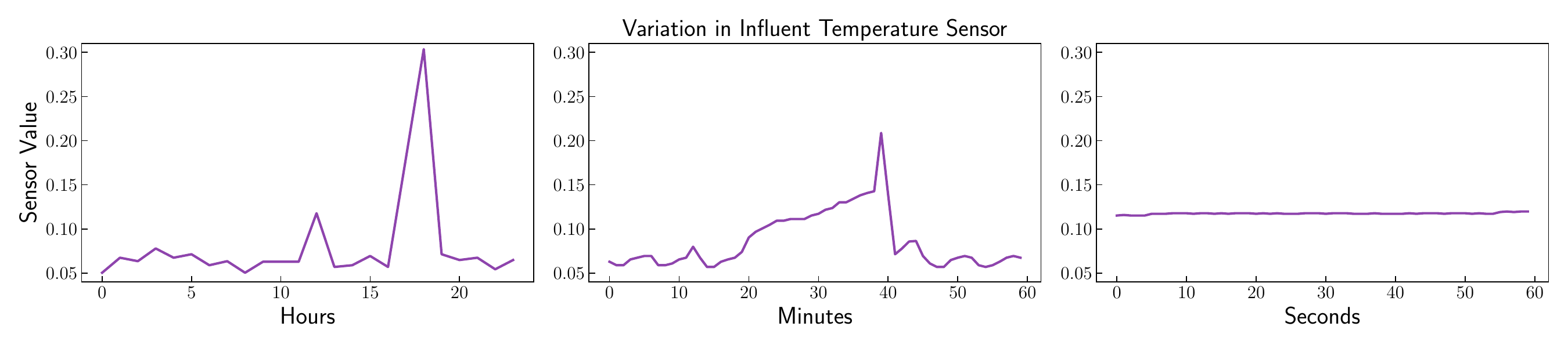}
	\caption{The many timescales of water treatment. Each subplot shows the incoming water temperature from the river at different temporal resolutions. Viewing left to right, if we look at temperature over the entire day (sub-sampled) we see a single outlier and an otherwise fluctuating baseline. In the middle subplot, looking at a single hour of data, we see the spike has more structure. Finally, the leftmost subplot shows one minute of data sampled at the fastest possible timescale of the system (no sub-sampling), which shows how in a short timescale measurements can even appear constant.} % On closer inspection the data does not smoothly change as suggested by the other subplots.}
	\label{fig:tit_3timescales}
\end{figure}

Our data exhibits a coherent structure over the year, month, day and minute. In Figure \ref{fig:tit_3timescales} we plot incoming water temperature at three temporal resolutions. Mechanical systems like ours often support sampling at rates of 1 Hz or greater, whereas data sets commonly used in time-series forecasting are wide and short; typically sampled once a day \footnote{Taking an extreme example, the well-known Sunspots dataset is unidimensional and contains 3240 data points.}. In water treatment, high-temporal resolutions are relevant because the data can be noisy (as highlighted in Figure~\ref{fig:mayclean}) and averaging is lossy. In addition, if one were to change process set-points (the ultimate end-goal of prediction), this may require rapid adjustment (for example, adjusting PID control parameters during a backwashing operation). 

%\begin{table}[!t]
%\centering
%\caption{A brief summary of different measurements each of the sensor type is responsible for working out.}
%\begin{tabular}{|l|l|l|}
%\hline
%\textbf{Sensor Type} & \textbf{Tag} & \textbf{Measures} \\ \hline
%Pressure & \textit{PIT} & Pressure exerted by pumps. \\ \hline
%Flowmeter & \textit{FIT} & Flow rate of the fluid. \\ \hline
%pH & \textit{PHIT} & Acidity and alkalinity of the solution. \\ \hline
%Temperature & \textit{TIT} & Temperature of the water. \\ \hline
%Turbidity & \textit{TUIT} & Turbidity of the water. \\ \hline
%Total Organic Carbon (TOC) & \textit{TCIT} & Organic carbon in the water. \\ \hline
%Conductivity & \textit{CIT} & Ability to pass an electric current. \\ \hline
%\end{tabular}
%\label{tab:summary}
%\end{table}

\subsection{Sudden, unpredictable events}
\begin{figure}[tb]
	\centering
	\includegraphics[width=\textwidth]{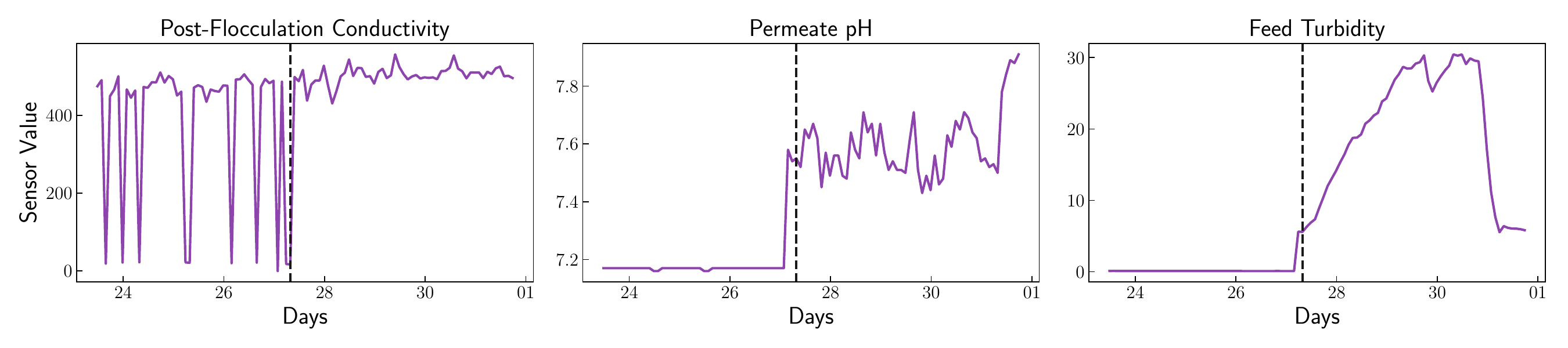}
	\caption{Raw values of some of the sensors before and after the cleaning. The black dotted line indicates when the sensors were manually cleaned by the plant operators. Note that the data is sub-sampled to avoid congestion in the plot. \looseness=-1}
	\label{fig:mayclean}
\end{figure}

Our data exhibit substantial distribution shifts, largely due to unpredictable events. For example, Figure \ref{fig:mayclean} shows the impact of cleaning different sensors. Most of these sensors get physically dirty over time due to a variety of factors. Sometimes water gets accumulated in the sensor enclosure, or moisture develops on the physical sensors, causing the readings to become noisy and unreliable. The plant operators manually clean the sensors to make sure they are as noise-free as possible and are reliably operating. Often times the sensor patterns indicate that they have recently undergone cleaning. This change in pattern manifests itself as the sensor signal stabilizes over time post-cleaning. 

\begin{figure}[!tb]
	\centering
	\includegraphics[width=\textwidth]{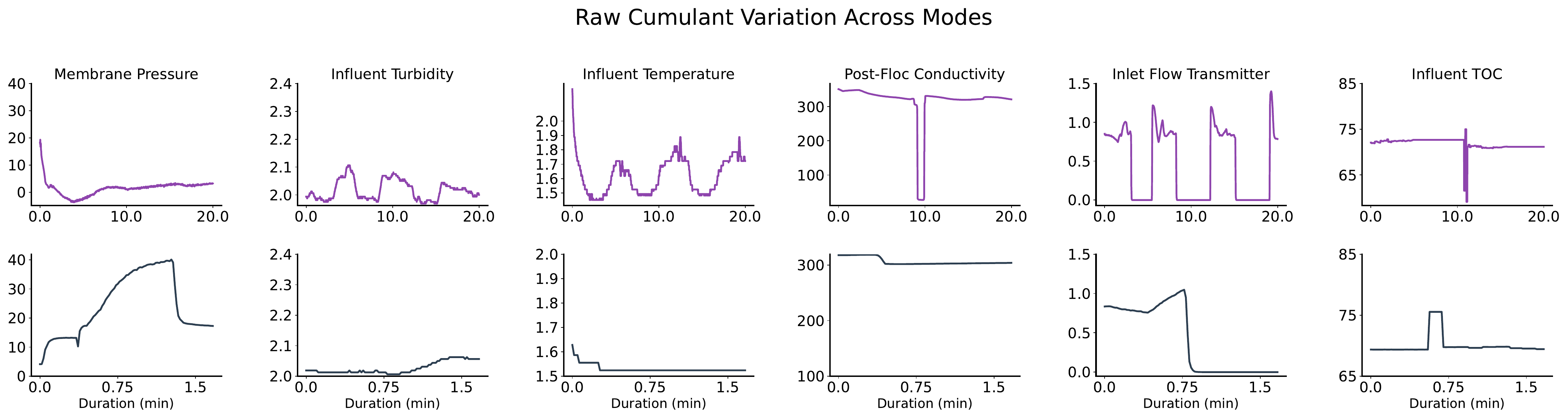}
	\caption{Variation across modes of different sensors. For brevity, we only produce two important modes, namely production (PROD), and backwashing (BW). The top row corresponds to the production mode, while the bottom row corresponds to the backwashing mode. The x-axis of the backwash data (second row) is plotted over a much shorter time scale because backwash only lasts for a couple of minutes, whereas production durations are much longer.}
	\label{fig:variation_modes}
\end{figure}

A water treatment plant operates in different modes which dramatically impacts the data generated. The main modes of operation are production and backwash. In production, the water is drawn through the filter to remove containments and it is moved to storage. In backwashing---the process of cleaning the filters---water moves backward through the system from storage, through the filters and eventually into the waste (reject) drain. In Figure  \ref{fig:variation_modes} we can see the impact of these two modes across several sensors. 

In our plant, mode change is driven either by a fixed schedule or human intervention. Maintenance, for example, occurs every day at 4:30 am, triggering the Membrane Integration Test (MIT) mode, whereas backwashing occurs on a strict schedule. Sensor changes due to these mode changes should be predictable from the time series itself, however, more ad hoc operator interventions are better represented as unpredictable external events; for example, when the plant is shut down. In addition, unscheduled maintenance occurs periodically---it is conceivable that such maintenance could be predicted based on the state of the plant, but there are other constraints like staffing constraints that can drive mode change. Later we discuss how we encoded the plant, which was key for successful prediction. 

\subsection{Sensor drift and seasonal change}
\begin{figure}[!htb]
	\centering
 	\includegraphics[width=\textwidth]{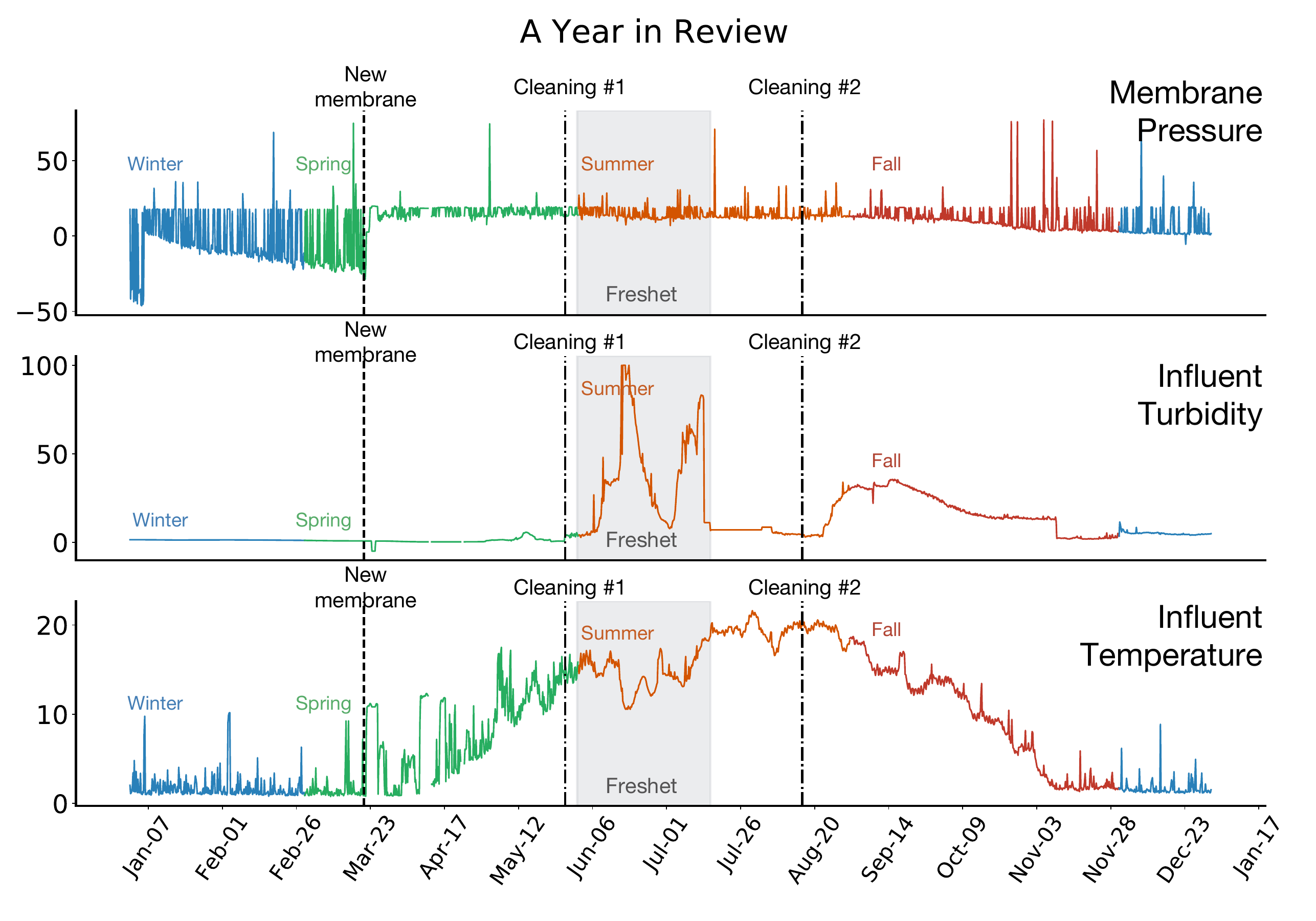}
	\caption{A year's worth of data for three different sensors.%, namely Membrane Pressure, Influent Turbidity, and Influent Temperature. 
	These three sensors are representative of the impacts that seasonal variations, or changes in the physical state of plant's components, have on the underlying telemetric stream of data. Note that the data is sub-sampled to avoid congestion in the plot. There are gaps in the data because the plant was down for maintenance.}
	\label{fig:yearlyreview}
\end{figure}

Water treatment is predominately driven by the conditions of incoming river water which changes throughout the year. These changes are driven by seasonal weather patterns. In the dead of Winter, the river is frozen and cool, clean, low turbidity water flows under the ice into the intake valves. During the Spring thaw---called the freshet---snow and ice all along the watershed of the river melt, increasing volume, flow, turbidity, and organic compounds in the river. Early Summer is dominated by a mixture of melted snow and ice higher up in the mountains and heavy rains that cause second and third freshets. Over the Summer, precipitation reduces, causing the late Summer and Fall to exhibit similar patterns as the Winter. All of these patterns are clearly visible in Figure~\ref{fig:yearlyreview}. 

\begin{figure}[tb]
	\centering
	\includegraphics[scale=0.40]{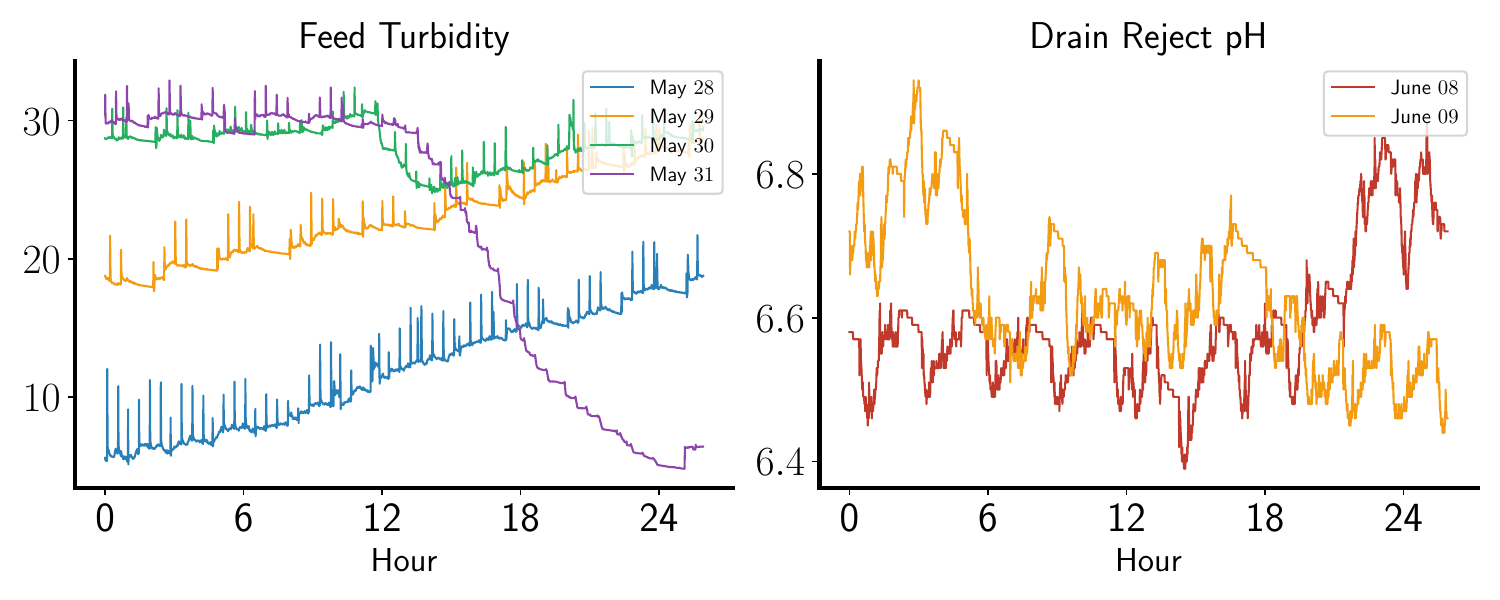}
	\caption{Feed Turbidity and Drain Reject pH sensors, respectively. An example of data drift in sensor values over hours. Each line represents a sensor plotted over a different day. We see variation in sensor readings both within a day and over multiple consecutive days.}
	\label{fig:drift_eg}
\end{figure}Change also happens within a single day. In Figure \ref{fig:drift_eg} we see how two different sensors evolve over a single day, on different days. As we can see in the plot of Feed Turbidity, some days are similar, but others, such as May~31, 2022, exhibit dramatically different dynamics. In some applications like HVAC control \cite{luo2022controlling}, it is sufficient to perform learning on a batch of data once a day. In water treatment, the sensor dynamics provide the opportunity to observe sensor changes throughout the day.

\subsection{The state of a water-treatment plant?}
What information would we need to predict water treatment data many steps into the future, with high accuracy? The plots above paint a clear picture of a partially observable complex dynamical system. Consider the Spring freshet. The volume and flow of the river will be driven by weather patterns and by the snow accumulation all along the watershed throughout the Winter. Digging deeper, the turbidity and other metrics are also driven by erosion and composition of the riverbed, which changes all the time. The chemical makeup of the water could spike if there is a change in farming practices in the area---water runoff from fields along the river. Even everyday things like a fire in the town can add huge pressure demands on the plant---many plants have dedicated pumps just for fires. 

In all the examples above, it would be impractical to sensorize these events so they could be detected in the plant. In fact, we would need to predict these events in advance of their occurrence (including the weather) in order to accurately predict our data in advance. Perhaps, we could simply make predictions based on the entire history of the time-series. The history would still only approximate the state, because we do not know the starting conditions: data from five and ten years ago. In addition, such an approach is not scalable if the end goal is to build a continual learning system that runs for years generating tens of millions of samples a year.

In the end, capturing the true underlying state is likely impossible and we must be content using learning methods that continue to learn in deployment in order to achieve accurate prediction. Such methods track the changing underlying state of the plant. The idea is to use computation and extra processing of the recent data to overcome the limitations of the agent's state representation~\cite{sutton2007role,tao2022agent}, similar to how an approximate model of the world can be used to deal with non-stationary tasks in reinforcement learning. In the next section, we will discuss different algorithms for learning and tracking in deployment and later show their advantages on water-treatment data.

\section{Multi-step Prediction}\label{sec:background}
In this paper we are interested in scalar predictions of multi-dimensional time-series, many steps into the future. On each discrete time-step, $t=1,2,...$, the learning algorithm observes a new observation vector, $\obs_t \in \mathbb{R}^d$, which form a sequence of vectors from the beginning of time.
\begin{equation*}\obs_{0:t} \doteq \obs_0, \obs_1, \obs_2, ..., \obs_t. \end{equation*}
We do not assume knowledge of the underlying process that generates the series. That is, the next generation of observation vector may depend not just on $\obs_{0:t}$, but also on other quantities not observable to the learning system. For example, the future turbidity of the river water is impacted by future weather which is not observable and generally not predictable.

The goal is to estimate some scalar function of the future values of the time-series on time-step $t$, given $\obs_{0:t}$.
In this paper, we focus on classical n-step predictions from time-series forecasting and exponentially weighted infinite horizon predictions commonly used in reinforcement learning, which we discuss in the following sections. 

\subsection{Classical Time-Series Forecasting}\label{sec_classical}
The first prediction problem we consider is simply predicting a component of the time-series on the next time-step, ${o}^{[i]}_{t+1}$. We denote the $i$-{\em th} component of $x_t$ as $x^{[i]}_t$. 
This scalar one-step prediction $\pred_t$ at time $t$ can be approximated as a function of a finite history of the time-series:
\begin{equation}\pred_t \doteq f_{\text{TS}}({o}^{[i]}_{t-\tau:t},\params_t)  \approx {o}^{[i]}_{t+1},\end{equation}
where $\params_t \in \mathbb{R}^k$ is the learned weights and $\tau$ is the number of previous observation vectors used to construct the prediction. For a classical autoregressive model, $f_{\text{TS}}$ is a linear function of this history ${o}^{[i]}_{t-\tau:t}$. More generally, $f_{\text{TS}}$ can be any nonlinear function, such as one learned by a neural network. 

In order to predict more than one step into the future we can iterate a one-step prediction model. The naive approach is to simply feed the model's prediction of the next observation into itself as input to predict the next step, now 2 steps into the future, and so on. For example a three-step prediction: 
\begin{equation}\pred_{t+2} \doteq f_{\text{TS}}([{o}^{[i]}_{t-\tau:t-1},\pred_{t+1}, \pred_{t}], \params_t) \approx {o}^{[i]}_{t+3}.\end{equation}
Notice how two components of the history of the time series have been replaced by estimates. As we iterate the model beyond $\tau$ steps into the future all the inputs to $f_{\text{TS}}$ will become model estimates.

Another approach is to directly learn an $n$-step prediction and avoid iterating altogether. 
One-step models are convenient because they can be updated at every timestep. Unfortunately, if the one-step model is inaccurate the model produces worse and worse predictions as you iterate it further. A {\em direct method} estimates an $n$-step prediction as a function of the history of the series:
\begin{equation}\pred_t \doteq f_{\text{DE}}({o}^{[i]}_{t-\tau:t},\params_t) \approx {o}^{[i]}_{t+p}.\end{equation}

In many applications, we are interested in multi-dimensional data and in predicting many steps in the future. We can go beyond auto-regressive approaches by simply considering these time series prediction problems as supervised learning problems. For example, we can learn a neural network $f_{\text{DE}}$ that inputs the last $\tau$ multi-dimensional observation vectors $\obs_{t-\tau:t}$ and predicts ${o}^{[i]}_{t+n}$, trained by constructing a dataset of pairs $(\obs_{t-\tau:t},{o}^{[i]}_{t+n})$. We can also go beyond finite $\tau$-length histories, and use recurrent neural networks, which is becoming a more common practice in time series prediction (see \citep[Section 2.3.1]{hewamalage2021recurrent}). When we start using this supervised learning framing, we lose some of the classical strategies for dealing with correlation in the data, but in general, evidence is mounting that we can obtain improved performance \cite{crone2011advances,hewamalage2021recurrent}.

\subsection{GVFs and Temporal Difference Learning}\label{sec_nexting}
\begin{figure}[!htb]
	\centering
	\includegraphics[width=\textwidth]{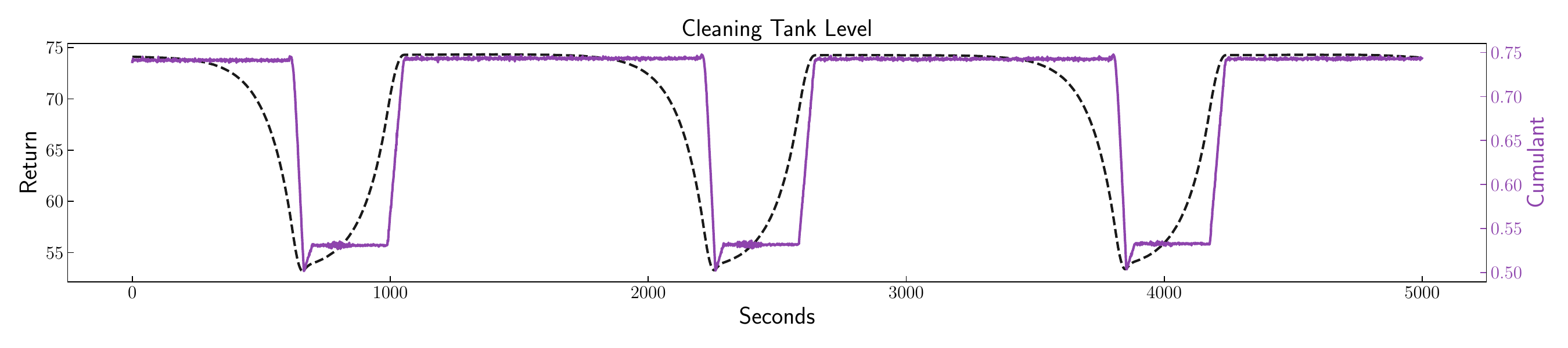}
	\caption{A sample time-series of tank level from a real water-treatment plant and an idealized prediction (labelled return).  The x-axis is the time-step, corresponding to one second. The prediction is ideal in the sense that we can simply compute the exponentially weighted sum in Equation \ref{eq:nexting} given a dataset---the idealized prediction is not the output of some estimation procedure. Later we will show learned predictions and how they match the ideal. Notice how the idealized prediction increases well before the time-series reaches its maximum value, and falls well before the time-series does. In this way, the idealized prediction at any point in time provides an anticipatory measure of the rise or fall of the data in the future. This is discussed extensively in prior work~\citep{modayil2014prediction}, so we do not belabour the point here.}
	\label{fig:nexting_eg}
\end{figure}

In reinforcement learning, multi-step predictions are formalized as value functions. Here the objective is to estimate the discounted sum of all the future values of some observable signal, with discount $\gamma\in[0,1):$
\begin{equation}G_t \doteq \sum_{j=0}^\infty \gamma^j {o}^{[i]}_{t+1+j}
. \label{eq:nexting} \end{equation}
Technically $G_t$ summarizes the infinite future of the time-series, but values of ${o}^{[i]}$ closer to time $t$ contribute most to the sum.\footnote{Note that we do not explicitly give the (partially observable) Markov decision process formalism because we do not need that precise notation to explain the concepts. Further, the predictions we consider are all on-policy predictions, so we do not need to know the explicit decision-making policy in order to do the update. For this reason, we avoid introducing all the notation around actions and policies, since they will not be used.} These exponentially weighted summaries of the future automatically smooth the underlying data ${o}^{[i]}$---potentially making estimation easier---and provide a continuous notion of anticipation of the future as discussed in Figure \ref{fig:nexting_eg}. For this reason, they have been called ``Nexting" predictions \citep{modayil2014prediction}, but more generally were introduced as general value functions (GVFs) \citep{sutton2011horde}, where they generalize the notion of a value by allowing any cumulant to be predicted beyond a reward. 

GVF predictions can be learned using temporal difference learning. As before, the prediction is approximated with a parameterized function, $f_{\text{TD}}(\svar_{t}, \params) \approx G_t$, where $\svar_{t}$ is a summary of the entire series, $\obs_{0:t}$, up to time $t$. For example, we could use an RNN to summarize this history; we opt for an even simpler approach---memory traces---which we describe in Section \ref{sec_state}. The prediction on time-step $t$ is updated using the temporal-difference error
\begin{equation}\params_{t+1} \leftarrow  \params_t + \alpha (c_{t+1} + \gamma f_{\text{TD}}(\svar_{t+1}, \params_t) - f_{\text{TD}}(\svar_{t}, \params_t))\nabla f_{\text{TD}}(\svar_{t}, \params_t), \label{eq:td} \end{equation}
where $\alpha\in(0,1]$ and $c_t \doteq {o}^{[i]}_t$.

\section{Methods}\label{sec:learnability}

In this paper, we investigate methods that can be pre-trained from offline logged data and perform fine-tuning in deployment. 
The algorithms we investigate can be used offline, online, or a combination of the two. Offline algorithms can randomly sub-sample and update from the offline data as much as needed (i.e. until the training loss converges). Online data, generated in the {\em deployment phase} can only be resampled from a replay buffer once it has been observed. Using the online data is restricted: the algorithms cannot look ahead into the future of the time-series, they must wait for each data point to become available step-by-step. After a sample is observed it can be resampled over and over via a replay buffer. In this section, we outline the algorithms, and how they can combine offline and online learning. 

\subsection{Constructing Agent-state}\label{sec_state}
These algorithms can be used in real-time, making and updating predictions live as the plant is operating. For simplicity in our experiments, we only simulate that setting here using a static offline dataset. The agent can iterate, in order, on the batch of offline data and it is equivalent to having made predictions live on the plant. 
%We consider 24 consecutive days of data as the offline logs of operator data, and the 25th day to represent our online data. The data from the 25th day is processed as a stream, one sample at a time, as if it were generated live. 
The offline batch of data is \(\mathcal{D}_{\text{offline}} = \{(o_{t}, c_{t+1}, o_{t+1})\}_{t=1}^N\), where \(N\) is the total number of transitions, \(o_{t}\in \mathbb{R}^d\) is the observation vector, \(c_{t+1}\in \mathbb{R}\) is the signal to predict  or {\em cumulant}, \(o_{t+1}\) is the next observation vector.

The data, however, is partially observable and the agent should construct an approximate state. A typical approach used in machine learning is to use RNNs, to summarize history \cite{hochreiter1997long,cho2014learning,hausknecht2015deep,vinyals2019grandmaster}. However, we found for our sensor-rich problem setting, that a simpler trace-based memory approach was just as effective and much easier to train. The general idea is to use an exponentially weighted moving average of the observations; such an exponential memory trace has previously been shown to be effective~\cite[c.f.][]{mozer1989focused,tao2022agent,rafiee2023from}. We include more explicit details on how we created our approximate state observation vector in Appendix \ref{app_state}.

Once we have constructed this approximate state vector, which we denote $\shat_t \in \mathbb{R}^{d+k}$, we then apply the algorithms directly on this $\shat_t$ without further considering history or state estimation.
% MARTHA: This is redundant with above. lets move it above, if we want to keep it? 
%These augmentations are sometimes referred to as auxiliary inputs and, in toy problems, it has been shown they alleviate the problem of partial observability in reinforcement learning~\cite{tao2022agent}. 
In other words, we construct an augmented dataset \(\mathcal{D}_{\text{augmented}} = \{(\shat_t, c_{t+1}, \shat_{t+1})_{t=1}^{N}\}\) and apply our algorithms as if we have access to the environment state---namely as if we are in the fully observable setting. 
All the algorithms we consider use a neural network $f$ to compute the prediction $f_{\params_t} (\shat_t)$,
where $\params_t$ are the parameters of the neural network. The predictions may either be GVF predictions or n-step time series predictions, with the algorithms described in the next two sections. 

\subsection{Algorithms for GVFs}

The goal for GVF predictions it to estimate the expected discounted sum of future cumulants, as described in Section \ref{sec_nexting}. 
%\subsection{OnlineTD: Online Temporal Difference Learning}
The simplest approach is to simply use the textbook 1-step temporal difference (TD) learning \cite{sutton2018reinforcement}. Data is processed as a stream, one sample at a time. The approach is summarized in Algorithm \ref{alg:tds}.
\begin{algorithm}[!htb]
\caption{OnlineTD}\label{alg:tds}
\begin{algorithmic}[1]
	\State Hyperparameters: stepsize $\eta > 0$
        \State Initialize \(\vect{w_{0}}\): the weights of the network (e.g., uniform)
        \State Obtain initial observation $o_t$ for $t = 0$,  set $\shat_0 = o_0$
        \While{\text{in deployment}}
        \State Observe next observation $o_{t+1}$ and cumulant $c_{t+1}$
         \State \(\shat_{t+1} \leftarrow U(o_{t+1}, \shat_t)\) \Comment{\textcolor{deepblue}{compute augmented observation vector}}
            \State \(v_{t+1} \leftarrow f_{\params_{t}}(\shat_{t+1})\) \Comment{\textcolor{deepblue}{compute prediction}}
            \State \(\delta_{t} \leftarrow c_{t+1} + \gamma v_{t+1} - f_{\params_{t}}(\shat_{t})\) \Comment{\textcolor{deepblue}{compute the TD error}}
            \State \(\params_{t+1} \gets \params_{t} + \eta \delta_{t}\nabla f_{\params_t}(\shat_{t})\) \Comment{\textcolor{deepblue}{or Adam using $-\delta_{t}\nabla f_{\params_t}$ as a gradient}}
            \State $t \gets t+1$
 	   % \State \(\shat_{t} \leftarrow \shat_{t+1}\)
        \EndWhile
    \end{algorithmic}
\end{algorithm}

We can also adapt this update to an offline dataset.
%\subsection{OfflineTD: Offline learning with TD}
We can use TD offline, making multiple passes over the data set, and updating the network weights via mini-batches. Here we follow the standard approach used in offline RL for the fully observable setting. In other words, we can treat each tuple $(\shat_{t}, c_{t+1}, \shat_{t+1})$ separately, without having to keep the data in order. In contrast, if we were using a recurrent neural network, we would need to maintain the dataset order more carefully. In each epoch, we shuffle the dataset \(\mathcal{D}_{\text{augmented}}\) and update the neural network using a mini-batch TD update. 
%construct $N/k$ non-overlapping consecutive mini-batches updating the neural network weights with each. 
Algorithm \ref{alg:tdroff} summarizes the approach. 

We use the Adam optimizer \citep{kingma2014adam} to update with the mini-batch TD updates. We set all but the stepsize $\eta$ to the typical default values: momentum parameter to $0.9$, exponential average parameter to $0.99$, and the small constant in the normalization to $10^{-4}$. The algorithm returns the state of the optimizer---such as the exponential averages of squared gradients and momentum---because our online variants continue optimizing online starting with this optimizer state. 

\begin{algorithm}[tb]
\caption{OfflineTD}\label{alg:tdroff}
\begin{algorithmic}[1]
	\State Hyperparameters: stepsize $\eta > 0$, batchsize $k$, number of epochs $n_{\text{epochs}}$, 
	\State Input \(\mathcal{D}_{\text{augmented}} = \{(\shat_{t}, c_{t+1}, \shat_{t+1})\}\) 
        \State Initialize \(\params\): the weights of the network (e.g., uniform)
        \State Initialize \(s_{\text{opt}}\): the state of the optimizer (e.g., zero momentum, zero exponential average)
            \For{epoch in \(n_{\text{epochs}}\)}
                \For {batch in \(\mathcal{D}_{\text{offline}}\)} 
                            %        \State \(\{(\shat_{t}, c_{t+1}, \shat_{t+1}, \gamma)_k\} \leftarrow \texttt{batch}\)
            \State $\Delta \leftarrow -\frac{1}{k}\sum_{i \in \text{batch}} \left( c_{i+1} + \gamma f_{\params}(\shat_{i+1}) - f_{\params}(\shat_{i}) \right)\nabla f_{\params}(\shat_{i})$
            \State \(\params, s_{\text{opt}} \gets \verb+opt+(\params , \Delta, \eta, s_{\text{opt}})\) 
                \EndFor
            \EndFor
            \State Return $\params, s_{\text{opt}}$
    \end{algorithmic}
\end{algorithm}

\begin{algorithm}[t]
\caption{OnlineTD using Offline Pretraining}\label{alg:tdo}
\begin{algorithmic}[1]
	\State Hyperparameters: offline stepsize $\eta > 0$, batchsize $k$, number of epochs $n_{\text{epochs}}$, online stepsize $\alpha > 0$, number of replay steps $n_{\text{replay}}$
	\State Input \(\mathcal{D}_{\text{augmented}} = \{(\shat_{t}, c_{t+1}, \shat_{t+1})\}\) 
	\State $\vect{w_{0}}, s_{\text{opt}} = $ OfflineTD($\mathcal{D}_{\text{augmented}}$, $\eta$, $k$, $n_{\text{epochs}}$)
	 \State Initialize the replay buffer \(\mathcal{B}\) with last $n_{\text{replay}}$ samples in $\mathcal{D}_{\text{augmented}}$
        \State Obtain initial observation $o_t$ for $t = 0$, set $\shat_0 = o_0$
        \While{\text{in deployment}}
        \State Observe next observation $o_{t+1}$ and cumulant $c_{t+1}$
         \State \(\shat_{t+1} \leftarrow U(o_{t+1}, \shat_t)\) \Comment{\textcolor{deepblue}{compute augmented observation vector}}
            \State \(v_{t+1} \leftarrow f_{\params_{t}}(\shat_{t+1})\) \Comment{\textcolor{deepblue}{compute prediction}}
            \State \(\delta_{t} \leftarrow c_{t+1} + \gamma v_{t+1} - f_{\params_{t}}(\shat_{t})\) \Comment{\textcolor{deepblue}{compute the TD error}}
             \State \(\params_{t+1}, s_{\text{opt}} \gets \verb+opt+(\params_t, -\delta_{t}\nabla f_{\params_t}(\shat_{t}), \alpha, s_{\text{opt}})\) 
          %  \State \(\params_{t+1} \gets \params_{t} + \alpha \delta_{t}\nabla f_{\params_t}(\shat_{t})\) 
            \State $t \gets t+1$
 	   % \State \(\shat_{t} \leftarrow \shat_{t+1}\)
        \EndWhile
    \end{algorithmic}
\end{algorithm}

%This method is largely included as a simple baseline. 
We expect purely offline methods to perform poorly in our non-stationary (partially observable) setting compared with those that also update in deployment. The offline data may not perfectly reflect what the agent will see in deployment, and, in general, tracking---namely updating with the most recent data---can also help under partial observability. 

We can combine the offline and online methods, by pre-training offline and then allowing the agent to continue learning online. 
The primary nuance here is that we can either continue to use a replay buffer to update online or switch to the simplest online variant of TD that simply updates once per sample. We found that the simpler update was typically just as good as the variant using replay, so we use this simpler variant in this work. We summarize this procedure in Algorithm \ref{alg:tdo}, and for completeness include the replay variant and results comparing to it in Appendix \ref{app_replay}.

It is worth mentioning that we could further improve these algorithms with the variety of advances combining TD and neural networks. TD methods can diverge when used with neural networks \cite{tsitsiklis1997analysis}, and several new algorithms have proposed gradient-based versions of TD that resolve the issue \cite{dai2017learning,dai2018sbeed,patterson2022generalized}. In control, a common addition is the use of target networks, which fix the bootstrap targets for several steps \cite{mnih2015humanlevel}. We found for our setting that the simple TD algorithm was effective, so we used this simpler approach.  
 
 \begin{algorithm}[tb]
\caption{OnlineNStep using Offline Pretraining}\label{alg:nstep}
\begin{algorithmic}[1]
	\State Hyperparameters: offline stepsize $\eta > 0$, batchsize $k$, number of epochs $n_{\text{epochs}}$, online stepsize $\alpha > 0$
	\State Input \(\mathcal{D}_{\text{n-step}} = \{(\shat_{t}, c_{t+n})\}\) 
	\State $\vect{w_{0}}, s_{\text{opt}} = $ OfflineNStep($\mathcal{D}_{\text{n-step}}$, $\eta$, $k$, $n_{\text{epochs}}$)
	 \State Create size $n$ circular array PastStates set index ind $\gets 0$ 
        \State Obtain initial observation $o_t$ for $t = 0$, set $\shat_0 \gets o_0$
         \State PastStates$[\text{ind}] \leftarrow \shat_{0}$, and $\text{ind} \gets 1$
           \For{$n-1$ steps} \Comment{\textcolor{deepblue}{store first $n$ inputs}}
        		\State Observe next observation $o_{t+1}$ and cumulant $c_{t+1}$
         	\State \(\shat_{t+1} \leftarrow U(o_{t+1}, \shat_t)\) 
         	\State PastStates$[\text{ind}] \leftarrow \shat_{t+1}$
	            \State $t \gets t+1$ and $\text{ind} \gets \text{ind} + 1$
	\EndFor
	\State $\text{ind} \gets 0$
        \While{\text{in deployment}}
        \State Observe next observation $o_{t+1}$ and cumulant $c_{t+1}$
         \State $(s, c) \gets (\text{PastStates}[\text{ind}],  c_{t+1})$
         \State $\Delta \gets  \left( f_{\params_t}(s) - c \right)\nabla f_{\params_t}(s)$
         \State $\params_{t+1}, s_{\text{opt}} \gets \verb+opt+(\params_t , \Delta, \alpha, s_{\text{opt}})$    
         \State \(\shat_{t+1} \leftarrow U(o_{t+1}, \shat_t)\) 
         \State PastStates$[\text{ind}] \leftarrow \shat_{t+1}$
          \State $t \gets t+1$ and $\text{ind} \leftarrow \text{mod}(\text{ind}, n)$
        \EndWhile
    \end{algorithmic}
\end{algorithm}

\subsection{Algorithms for n-step Predictions}
\label{sec:nstep}

We can similarly consider the offline and online variants of n-step predictions. The offline dataset\footnote{The underlying data is the same as in the TD setting, but the targets are different, and so we explicitly construct a supervised learning dataset from this underlying data.} consists instead of \(\mathcal{D}_{\text{n-step}} = \{(\shat_{t}, c_{t+n})\}_{t=0}^{N-n}\) where we predict the cumulant $n$ steps into the future from $t$, given the approximate state $\shat_{t}$. The targets for GVF predictions were returns $G_t$---discounted sums of cumulants into the future---whereas the targets for n-step predictions are the cumulants exactly $n$ steps in the future. Learning $f_{\params}$ offline corresponds to a regression problem on this dataset, which can be solved using any standard techniques. Similarly to OfflineTD, we use stochastic mini-batch gradient descent and the Adam optimizer. 

As a supervised learning problem, it is straightforward to update in deployment, online. However, there is one interesting nuance here, that the targets are not observed until $n$ steps into the future. The online algorithm, therefore, has to \emph{wait} to update the prediction $f_{\params}(\shat_{t})$ until it sees the outcome $c_{t+n}$ at time step $t+n$. This involves maintaining a short buffer of size $n$, until we can obtain the pair $(\shat_{t}, c_{t+n})$. This procedure is summarized in Algorithm~\ref{alg:nstep}. 

\begin{figure}[htb]
	\centering
	\includegraphics[width=\textwidth]{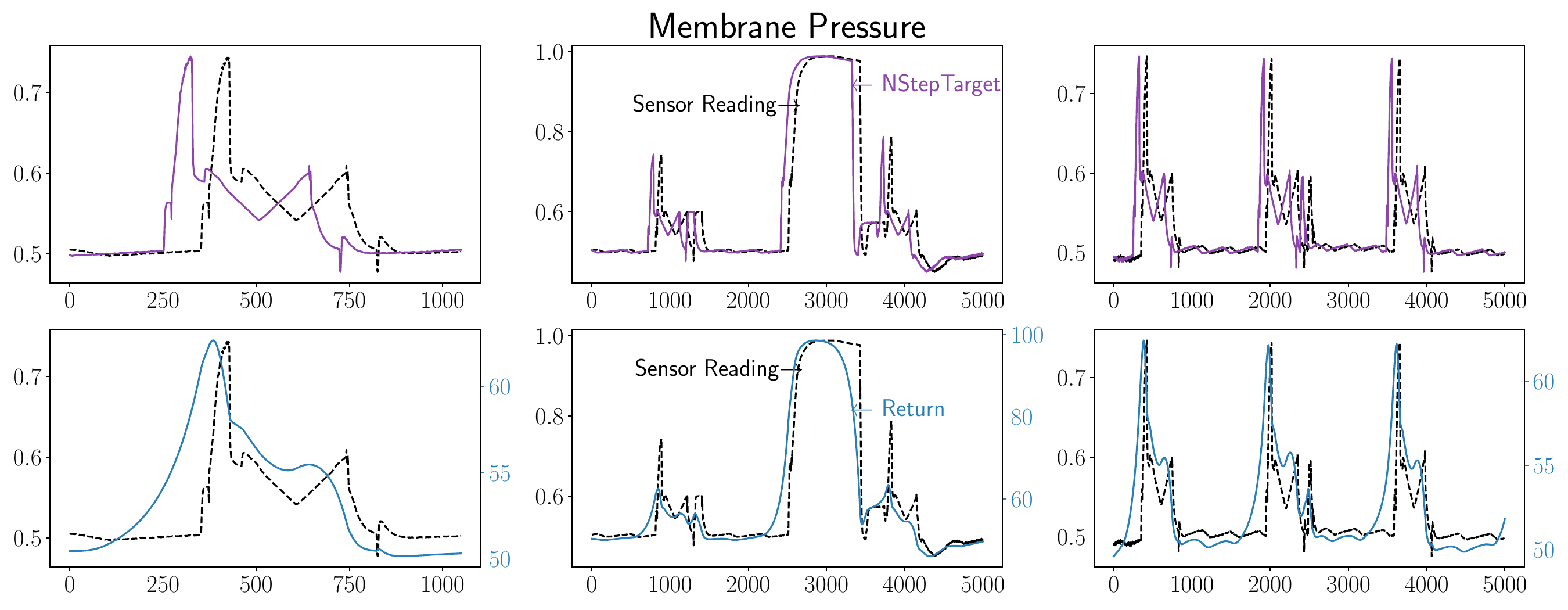}
	\caption{{\bf Samples of Membrane Pressure sensor time-series and the corresponding idealized GVF and n-step predictions.} Consider the first row of subplots. Each subplot shows a different snippet of time: the dashed black line depicts the time-series of Membrane Pressure and the purple line shows the corresponding prediction target or ideal prediction. Since the purple line is an ideal prediction of the black line, the purple line is shifted forward in time (to the left) revealing events before they occur in the dashed black time-series. The second row, similar to the first, compares the pressure sensor reading time-series and its ideal prediction for TD (labelled ``return").   }
	\label{fig:pred_targets}
\end{figure}

Though seemingly a minor issue, it is less ideal that the OnlineNStep algorithm has to wait $n$ steps to update the prediction for input $\shat_{t}$. The TD algorithm for GVF predictions, on the other hand, does not have to wait to update, because it bootstraps off of its own estimates. Instead of using bootstrapping, we could have used a Monte Carlo algorithm, that regresses $\shat_{t}$ towards computed returns, turning this into a supervised learning problem like for the n-step time series problem. However, it has been shown that being able to update immediately can result in faster tracking \citep{sutton2007role,sutton2018reinforcement}, and typically TD algorithms are preferred to Monte Carlo algorithms. The issue is worse for Monte Carlo than for n-step targets, because the returns extend further than $n$ steps into the future, but nonetheless, there is some suggestive evidence that algorithms that need to wait could be disadvantaged.

N-step and GVF predictions are quite similar in the sense that their fundamental role is to summarize the future of a time-series, which is easy to see when looking at real data. Figure \ref{fig:pred_targets} plots the prediction targets for n-step predictions and GVF predictions (learned by TD) on real sensor data. 

%We can also use classic time series prediction approaches for the N-step targets, as discussed in Section \ref{sec_classical}. 
One might wonder why we chose to use the same agent-state construction and neural network for the N-step targets, as for the GVF targets, when there are many time series prediction approaches to chose from. Our primary reason is that we found this supervised approach to be effective, in terms of forecasting accuracy. This finding actually well-matches recent analysis, that highlights that for larger, multivariate time series data, neural network approaches can be more effective than the simpler time-series approaches \citep{hewamalage2021recurrent}. Essentially, the nonlinear modeling power of neural networks becomes useful in these bigger data regimes, whereas the simpler methods remain preferable for the typically smaller datasets in the time series literature. We did test a time-series approach called NLinear that has been recently shown to be competitive with state-of-the-art prediction methods, including methods based on transformers~\cite{zeng2022transformers,zhang2023crossformer}. The performance was worse than our proposed approach for the N-step predictions, as we discuss in Section \ref{sec_experiments}.

\subsection{Selecting hyperparameters for deployment}\label{sec_hypers}

The above algorithms have many hyperparameters. Fortunately, we can use a simple validation strategy to select them, including the \emph{online} stepsize parameters. The key idea is to treat the validation just like a deployment scenario, where the agent updates in temporal order on the dataset. For example, consider selecting the offline stepsize $\eta$ and online stepsize $\alpha$, assuming all other hyperparameters are specified (number of offline epochs is fixed, etc.). Then we can evaluate each hyperparameter pair $(\eta, \alpha)$ by
\begin{enumerate}
\item splitting the dataset into a training and validation set,
\item pre-training with offline stepsize $\eta$ on the training set,
\item updating online with the stepsize $\alpha$ on the validation set (in one pass) as if it is streaming, recording the prediction accuracy as the agent updates.
\end{enumerate}

The online prediction accuracy is computed as follows. For the current weights $\params_t$, the agent gets $\shat_t$ and makes a prediction $\pred_t =  f_{\params_t}(\shat_{t})$. Because we (the experimenter) can peek ahead in the validation set, we can compute the error $\text{err}_t = (\pred_t - c_{t+n})^2$. The agent, of course, cannot peek ahead, since it would not be able to do so in deployment. After going through the validation set once, we have our set of errors. Note that we only evaluate $\params_t$ on the pair $(\shat_{t}, c_{t+n})$. Right after this step, we update the weights to get $\params_{t+1}$ and then evaluate the prediction under these new weights for the next step: $\pred_{t+1} =  f_{\params_{t+1}}(\shat_{t+1})$.

This validation procedure helps us pick a suitable pair of $(\eta, \alpha)$ precisely because validation mimics deployment. We want $\eta$ to be chosen to produce a good initialization and we want $\alpha$ to be chosen to facilitate tracking when updating online. For example, if $\alpha$ is too big for tracking (or fine-tuning), then the validation error will be poor because the weights will move away from a good solution while updating on the validation set and the errors will start to get larger, resulting in a poor final average validation error. As another example, if $\eta$ is too small and does not converge on the training set within the given number of epochs, then the initialization will not be as good and the validation errors will start higher than they otherwise could, until the online updating starts to reduce them. 

Though this hyperparameter selection approach is described specifically for n-step predictions with the offline and online stepsizes, it can be used for TD as well as for other hyperparameters. The key point is that, even though the offline hyperparameters are only used on the training set and the online hyperparameters only when updating on the validation, they are both jointly evaluated based on validation error. The primary difference for TD is simply that the target is different. Again, because we the experimenter can look ahead in the data, we can simply compute the return on the future data, and compute the errors $\text{err}_t = (f_{\params_t}(\shat_{t}) - G_{t})^2$.

\section{Experimental Setup}
\label{sec:expsetup}

We investigate a scenario where the agent pre-trains on offline data and it's prediction accuracy is then tested in deployment. The agent makes predictions on every time step in deployment, and we can retroactively check the accuracy of those predictions once we see the future---either after N steps or after enough steps to compute the return. For our experiments, we simply collect a test set and then have the agent predict incrementally on this test set, as if it is in deployment. This is perfectly equivalent to making predictions on the real system, but importantly allows us to run different algorithms on the same deployment (test) data. Further, it allows us to take our year of collected data, and select different time periods to split into train and test.

We consider two different scenarios: learning on 4 days, testing on 1 day and learning on 30 days, testing on 7 days. Most of our experiments consist of a dataset of five consecutive days of data from the middle of November 2022. The first four days are used as the offline training logs while the fifth day is used for the deployment phase.
We use the final 4k steps of 
 %We use the final \(2\%\) of 
 the offline training logs as the validation data, which is used for selecting hyper-parameters. 
 We also re-run all of experiments on another time period of five consecutive days in May---chosen because water conditions will be notably different from November---to ensure conclusions are not specific to November data.

 For the final experiment, we use data from the duration of an entire month: the data from the entire month of November is used as the offline training logs and the next week (December 1st to December 7th) is used as the deployment data. The final 7 days of the offline training logs are used as the validation set. In addition, we subsample, so that the timescale of sensors readings is every 10 seconds, rather than every second. The goal of this final experiment is to test the agent in a setting where deployment is further from training, likely making it more important to update during deployment. Note also that this final setting is more challenging, because the time horizon itself is further: a 100-step prediction for the five day data is 100 seconds in the future, whereas it is 1000 seconds for this final experiment. 

 % by selecting 1 second for every 10 seconds of data.
 
All the methods share similar settings. We train a 2-layer feed-forward neural network with 512 units in each layer with ReLU activation functions. The input to the network is an augmented observation vector of length 384, which is constructed by concatenating the 185 sensor values from the plant\footnote{The raw sensor vector was length 480. We removed all constant sensor readings, leaving 185 sensors.} with 185 memory traces and an additional 14 inputs encoding the plant mode (this is described in detail in Section \ref{sec:learnability} and Appendix A). In total, the network contains 722,945 weights. In the experiments that use five days of data, this network is optimized for 4000 epochs using the Adam optimizer with an L2 weight decay rate of $\lambda=0.003$ and a batch size of 512, in the offline phase.  After the offline training phase ends, we save the optimizer state variables and use them to initialize the optimizer during the deployment phase. In deployment, the algorithms update using one sample at a time, and use a different online step-size.
% MARTHAC: This is clear from the pseudocode
 %Note that in the case of OnlineTD and OnlineNStep, updates are done using only the next observations as opposed to a batch of observations.

For all the methods we use the validation procedure described in Section \ref{sec_hypers} to select the step-size parameter.
%For all the methods under comparison, we perform an exhaustive sweep over the different hyper-parameter values and pick the best setting based on the performance on the validation data.
 We swept over offline learning rates $\eta\in \{1 \textsf{x} 10^{-3},1 \textsf{x} 10^{-4}, 1 \textsf{x} 10^{-5}, 1 \textsf{x} 10^{-6}, 1 \textsf{x} 10^{-7}\}$ and online learning rates $\alpha \in \{1 \textsf{x} 10^{-4}, 1 \textsf{x} 10^{-5}, 1 \textsf{x} 10^{-6}, 1 \textsf{x} 10^{-7}, 1 \textsf{x} 10^{-8}\}$. The validation procedure is done separately for each algorithm and sensor.
 %, as the to select effective offline and online stepsizes 
 %Note that this hyper-parameter search is done separately for each algorithm and sensor.
 %A summary of the hyper-parameters is given in Appendix \ref{apx:algdetails}.

% 
% Although the WTP data is non-stationary, the impact of this non-stationarity is not very significant within the period of five days.
% Due to this, the training and deployment data are very similar, and online adaptation during the deployment phase is unlikely to have a significant impact on the final performance.
% Since the distribution shifts and pattern changes occur at larger time intervals, we prepare an additional dataset to highlight the necessity of online adaptation.
% For this purpose, the training phase uses the data from the duration of an entire month: the data from first 24 days of November are used as the offline training logs and the last day of the month (30th November) is used as the deployment data.
% In order to reduce the size of this dataset, we sub-sample by selecting 1 second for every 10 seconds of data.
% All the hyper-parameters remain the same for both datasets.
% However, we sweep over the learning rates $(\eta, \alpha)$ separately for each dataset.

\section{Experiments and Results}\label{sec_experiments}
A natural first question is can we predict the time series well in deployment, given the size, complexity, and partially observable characteristics of our data. From there we contrast the GVF predictions to n-step predictions, to better understand the GVF results relative to a well-understood multistep prediction. Finally, we investigate one of the key claims in this work: does learning in deployment help or is offline learning all we need?
% better ground this result relative to a well-understood multistep prediction. 
%there are many many possible subquestions. Did our (1) feature engineering (mode timers, time encoding, and traces) and learning online in deployment improve performance---potentially mitigating partial observability? Does pre-training on logs of data improve or aid the later learning in deployment phase? Finally, is learning in deployment needed---is offline learning all we need?

\begin{figure}[htb]
	\centering
	\includegraphics[width=\textwidth]{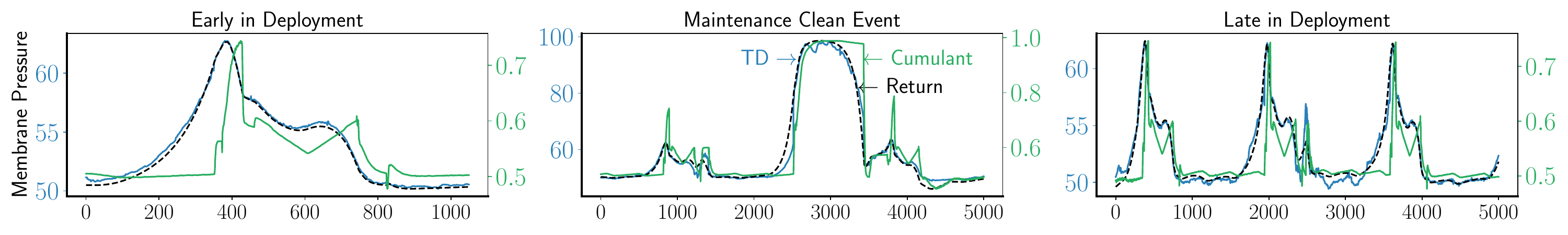}
	\caption{Predictions of the filter membrane pressure roughly 100 seconds into the future. The plot shows the pressure sensor in green labelled cumulant (whose magnitude corresponds to the right y-axis). We show three snippets of the deployment data. The first subplot shows (on the x-axis) a thousand time steps (seconds) at the beginning of deployment. The middle subplot shows data during a maintenance clean, and the last subplot features data near the end of the deployment phase (24 hours later). Each subplot highlights a different characteristic pattern in pressure change. The blue curve shows the TD prediction, first trained offline, then updated in deployment. The return represents the ideal prediction and is plotted in black. Note both the TD prediction and the return use the left blue axis. The TD predictions tightly match the target's pattern in all three scenarios.}
	\label{exp:result1}
\end{figure}

\paragraph{GVF predictions are accurate in deployment}
 The object of our first set of results is to gain some intuition about GVF predictions. Although widely used in RL to model the utility or value of a policy, exponentially weighted predictions are uncommon. In Figure \ref{exp:result1} we visualize predictions from the OnlineTD approach\footnote{All of our results are with pre-training, as this performed significantly better than without using the offline data at all. This result is to be expected. Furthermore, our OnlineTD algorithm with pre-training also leverages the offline data to automatically set all hyperparameters, providing a fully specified algorithm. The conclusion for our setting is that it simply makes the most sense to leverage offline data, rather than learning from scratch.} on one sensor at three different periods of time in deployment. Here we plot the cumulant (sensor value to be predicted into the future), the prediction, and the return---our stand-in for an {\em idealized prediction}. The time series of the return changes before the cumulant, because the return summarizes the future values of the cumulant. A good prediction should closely match the return as we see in the figure.

In the middle subplot of Figure \ref{exp:result1} we see a large perturbation in the cumulant corresponding to a difficult to predict event. This event, a maintenance clean, happens in the early morning. This causes a large increase in pressure on the filter, and unlike the vast majority of the training data, this increase is sustained for a long period of time. We can see the prediction correctly anticipates this event but does not get the precise shape of the prediction correct.

The particular time of year had a minimal impact on the quality of predictions learned. Figure \ref{exp:result1_may_data}, in the appendix is a replication of Figure \ref{exp:result1} with different training and deployment data, but the same sensor. We used the same architecture, preprocessing, and training scheme as described in the previous section and we see the predictions closely match the return as before.  

\begin{figure}[htb]
	\centering
	\includegraphics[scale=0.25]{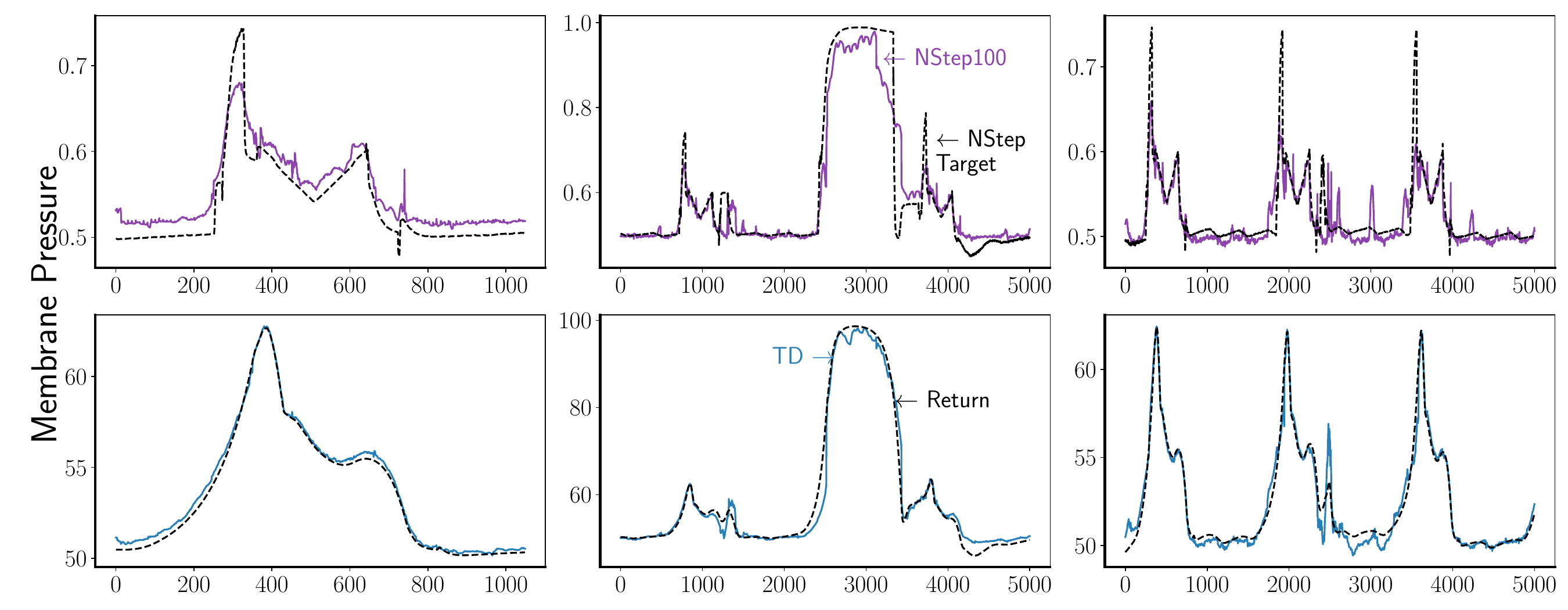}
	\caption{Comparing GVF Predictions (blue) and n-step predictions (purple) of filter membrane pressure. The top row shows the n-step predictions on the same three segments of deployment data used in Figure \ref{exp:result1}. As before, the x-axis is time-steps or seconds. Here we only plot the prediction (labelled TD and NStep100), and the ideal prediction (labelled return and NStep target). Although both types of predictions are well aligned with their respective targets, however, sometimes the n-step prediction is off. Figure \ref{exp:nstepVnexting} includes the results for several other sensors.}
	\label{exp:nstep}
\end{figure}
	
\paragraph{Comparing GVF and n-step predictions} 

To the uninitiated, GVF predictions can seem somewhat alien. To help calibrate our performance expectations, and provide a point of comparison, we also learned and plotted the more conventional 100-step predictions of future membrane pressure in deployment in Figure \ref{exp:nstep}. We chose a horizon of 100 steps to provide rough alignment with the horizon of a $\gamma=0.99$ GVF prediction. The horizon for a GVF prediction is typically said to be about $\tfrac{1}{1-\gamma}$ \citep{sutton2011horde}. The figure shows the n-step prediction and the GVF prediction on the same segments of data in deployment.  

The plot of the n-step prediction and the shifted cumulant (labelled NStep Target) should align if the predictions are accurate. At least for membrane pressure, the GVF predictions better match their prediction target (the return) compared with n-step predictions.
 \begin{figure}[htb]
	\centering
	\includegraphics[scale=0.2]{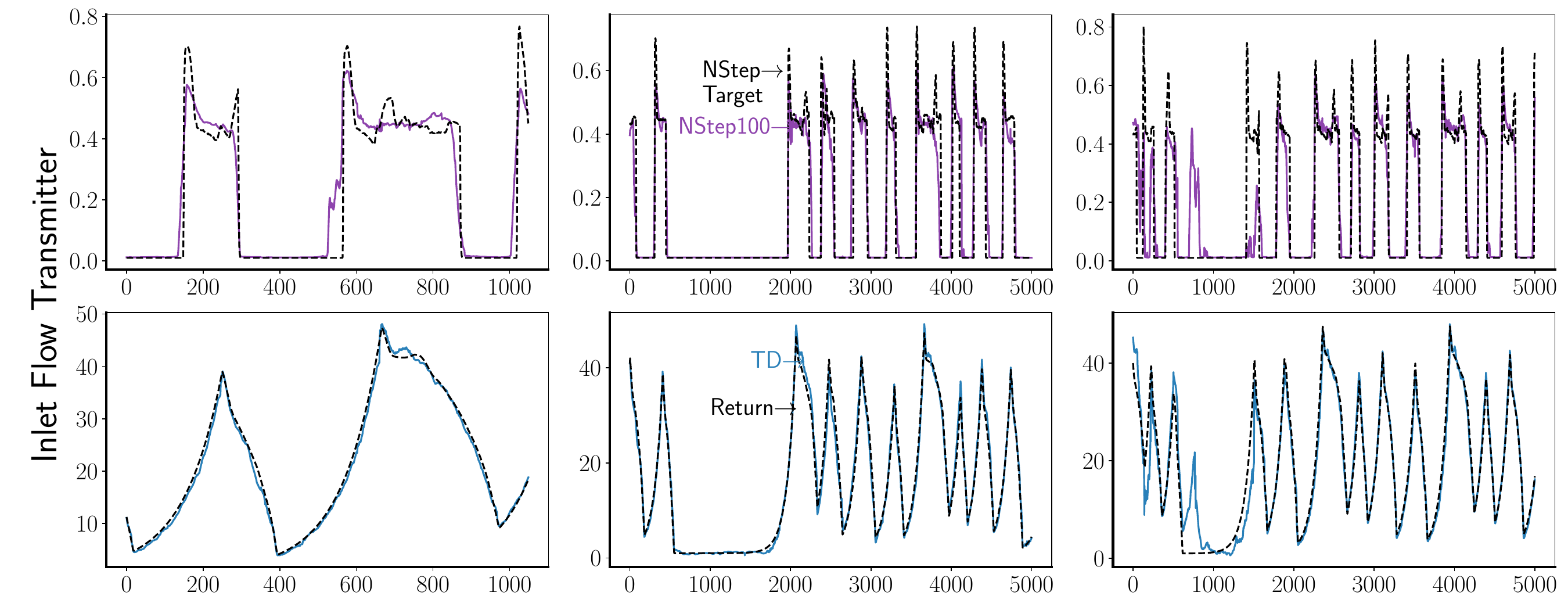}
	\includegraphics[scale=0.2]{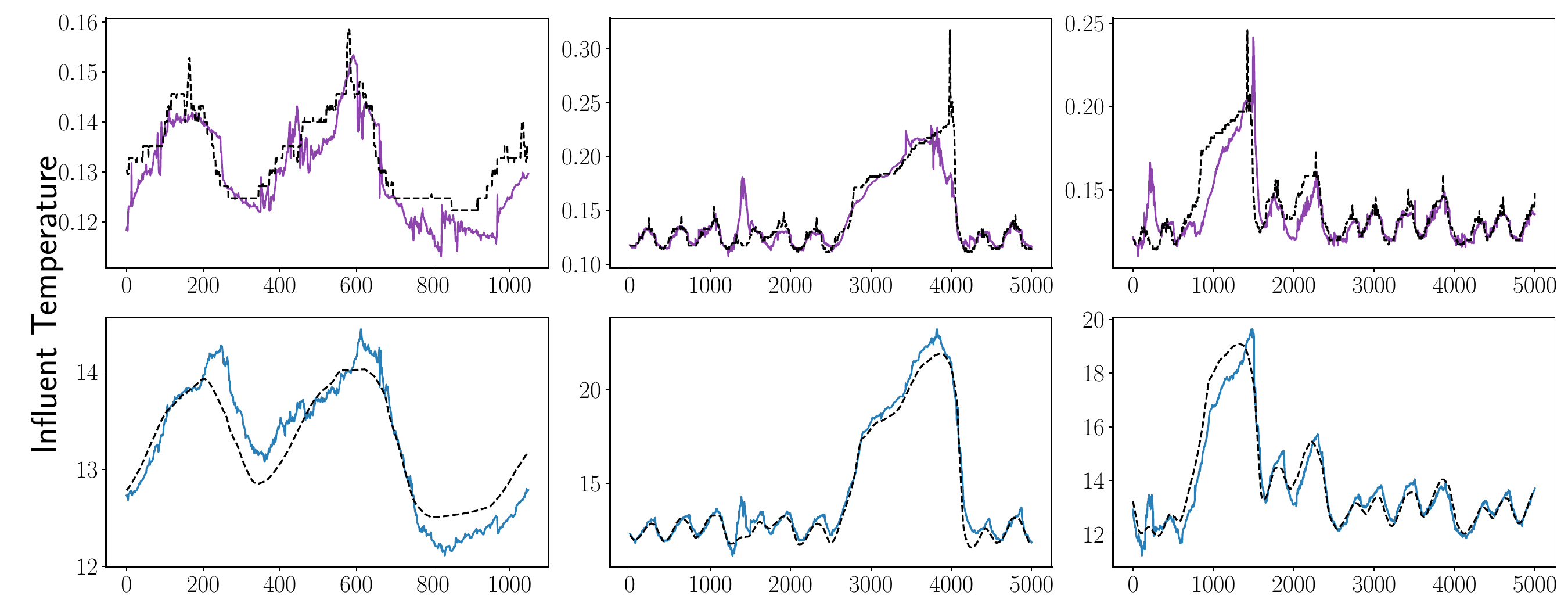}
	\includegraphics[scale=0.2]{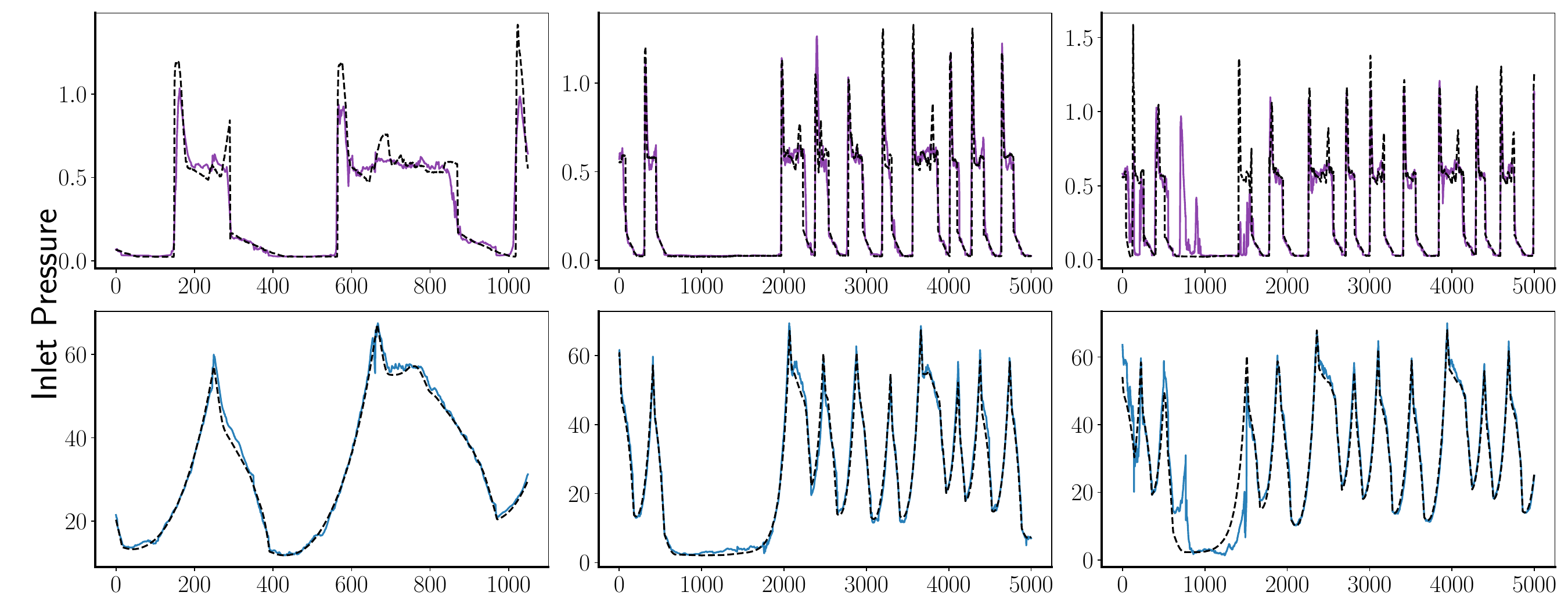}
	\includegraphics[scale=0.2]{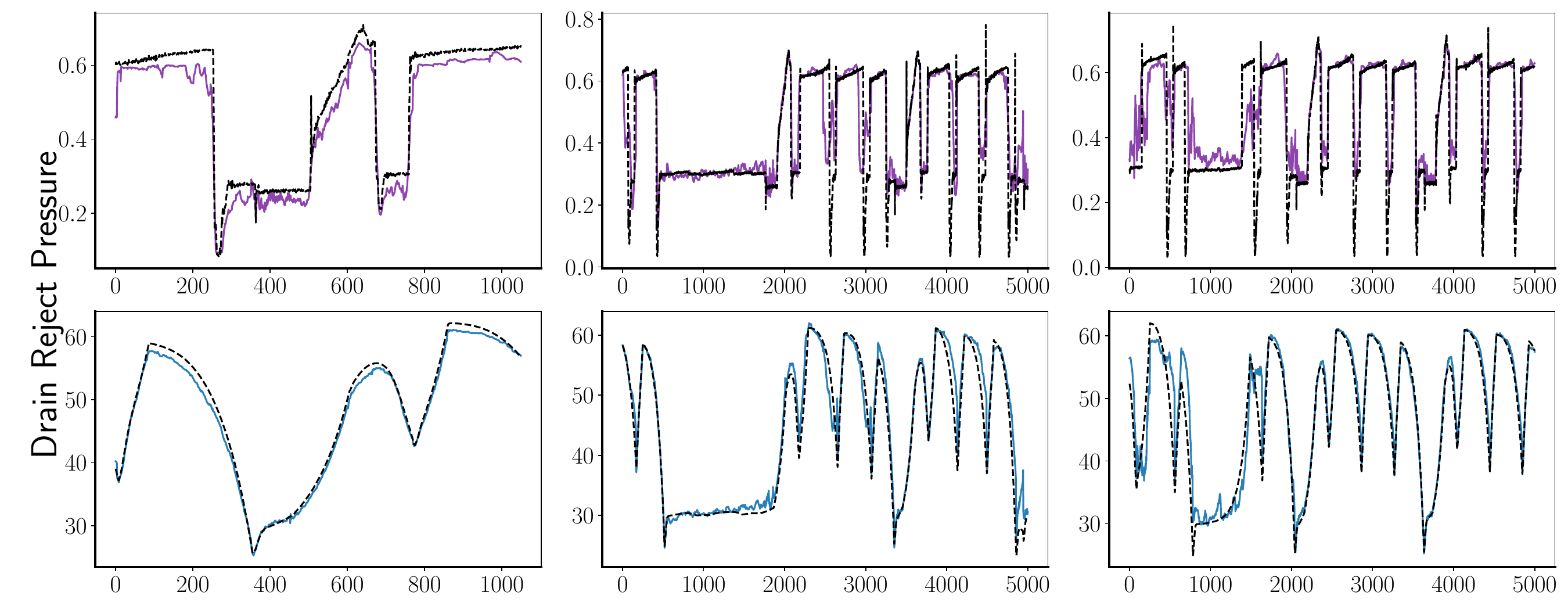}
	\caption{Comparing n-step predictions (Row 1) and GVF Predictions (Row 2) across several sensors. The structure of this plot mirrors Figure \ref{exp:nstep}.   }
	\label{exp:nstepVnexting}
\end{figure}

To get a better sense of the quality of these learned predictions, we also compared against a simple linear baseline commonly used in time-series forecasting. This method called NLinear has been recently shown to be competitive with state-of-the-art prediction methods, including methods based on transformers~\cite{zeng2022transformers,zhang2023crossformer}. NLinear simply learns a linear map from a history of the normalized sensor values to the n-step future target. We experimented with a short history length (336) closer to the length of the prediction horizon (n=100), and a much longer history (4000). The much longer history performed better, but generally replicated the periodicity of the sequence in its future predictions, overall leading to much worse performance compared to our n-step baseline. The results can be found in Figure \ref{exp:NLinear} in the Appendix. 
 
Generally, across sensors, the learned GVF predictions are smoother than their n-step counterparts as shown in Figure \ref{exp:nstepVnexting}. This is perhaps to be expected because the $\gamma$ weighting in the GVF prediction targets smooths the raw sensor data. If there are sharp, one-time-step spikes, as we see in the Inlet Pressure date, the n-step target itself will be spikey---that is, the ideal prediction is not smooth. Otherwise, the main objective of Figure \ref{exp:nstepVnexting} is to allow you the reader to better understand GVF predictions by simply visually comparing them with n-step predictions---something that is easy to interpret and you might have more natural intuitions for.

\begin{figure}[htb]
	\centering
	\includegraphics[width=0.5\textwidth]{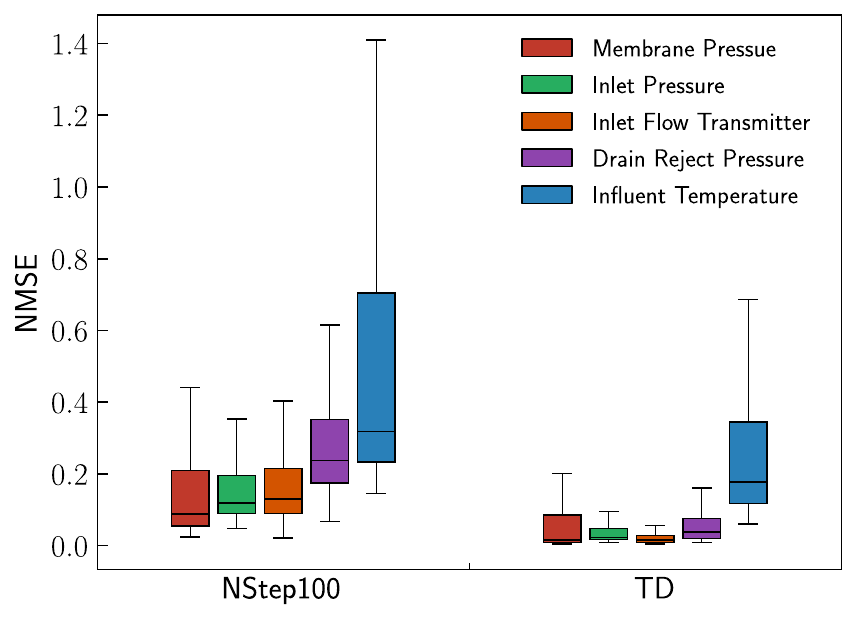}
	\caption{Standard box plots generated using the sequences of Normalized MSE observed throughout the entirety of the deployment data for GVF and n-step predictions on 5 different sensors. The horizontal line in the middle denotes the median value. The top and bottom box boundaries represent the 75th and 25th percentiles, respectively. The whisker boundaries are drawn at the data point that is located closest to the distance of 1.5 times the Interquartile Range.  Note that the NMSE values for TD and NStep100 are not directly comparable since their prediction targets are different. However, in both cases, it is desirable to have a lower NMSE value. The outliers are not plotted in order to make the visualization easier.}
	\label{fig:nstepVnexting}
\end{figure}

One perhaps surprising conclusion from Figure \ref{exp:nstepVnexting} is that the GVF and n-step predictions look surprisingly similar, and thus it is reasonable to ask if there are reasons to prefer one to the other. From a performance perspective, we compare the two in Figure \ref{fig:nstepVnexting} reporting the Normalized Mean Squared Error (NMSE) over the deployment data:
\begin{equation*}
{\text{NMSE}}_t \doteq \frac{{\text{MSE}}_t}{\overline{\sigma^2(G_t)}}\quad\quad\quad
\text{where} \quad {\text{MSE}}_t \doteq \overline{(\hat{v}_t - G_t)^2}
\end{equation*}
and $\overline{x}$ denotes the exponentially weighted moving average of the squared GVF prediction error over the deployment data. Similarly, $\overline{\sigma^2}(G_t)$ denotes the variance of the returns up to time $t$ computed using an exponentially weighted variant of Welford's online algorithm \citep{welford1962note}. The NMSE is equivalent to the variance unexplained and is a simple ratio measure of the MSE of the predictor to the MSE of the mean prediction. NMSE less than one indicates that the prediction explains more variance in the data than a mean prediction. We use the exponential moving average variants of these measures because our data is non-stationary. Finally, NMSE for the n-step prediction can be computed by replacing  $G_t$ with $o^{[i]}_{t+100}$ in the above equations. 
Across five sensors, the NMSE is lower for GVF predictions compared with n-step predictions, as shown in Figure~\ref{fig:nstepVnexting}.
%
%\begin{table}[htb]
%\centering
%\caption{Normalized MSE averaged over deployment for GVF and n-step predictions on five different sensors.}
%\label{tab:nstepVnexting}
%\begin{tabular}{|l|l|l|l|l|l|}
%\hline
%& \begin{tabular}[c]{@{}l@{}}Membrane \\ Pressure\end{tabular} & \begin{tabular}[c]{@{}l@{}}Influent\\ Temperature\end{tabular} & \begin{tabular}[c]{@{}l@{}}Inlet\\ Pressure\end{tabular} & \begin{tabular}[c]{@{}l@{}}Inlet Flow\\ Transmitter\end{tabular} & \begin{tabular}[c]{@{}l@{}}Drain Reject\\ Pressure\end{tabular} \\ \hline
%TD & 0.129 & 0.313 & 0.047 & 0.030 & 0.060 \\ \hline
%NStep100 & 0.212 & 0.768 & 0.182 & 0.190 & 0.306 \\ \hline
%\end{tabular}
%\end{table}

Algorithmically, GVF predictions are interesting for several reasons. GVF predictions can be updated, via TD, online and incrementally from a stream of data, whereas n-step predictions involve storing the data and waiting 100 steps until the prediction targets are observed. The longer the prediction horizon, the longer the system must wait without updating the predictions in between.
% MARTHA: The memory requirements are just so low...
% and the larger the memory footprint of the prediction methods. 
In contrast, TD methods by their recursive construction have memory and computational requirements independent of the prediction horizon---independent of 
$\gamma$. These points highlight the potential of GVF predictions for time-series prediction, as an additional choice for multistep predictions. For any given application, the ultimate choice of prediction type and learning method will be driven by many factors.

\paragraph{Mitigating partial observability via online adaption}
As shown extensively in Section \ref{sec:data_description}, the data from our plant is highly partially observable, appearing non-stationary when plotted. For the smaller dataset we used in the experiments so far, however, we find that the agent trained only on offline data predicts the deployment data well. This outcome is not surprising because the training data was collected from only four days of operation and the deployment was the following 24-hour period. It is reasonable to expect that the data is mostly stationary during this period since there would be no major seasonal weather changes, unexpected events like fires are rare, and sensor fouling takes weeks to show up in the data stream.

\begin{figure}[htb]
	\centering
	\includegraphics[width=\textwidth]{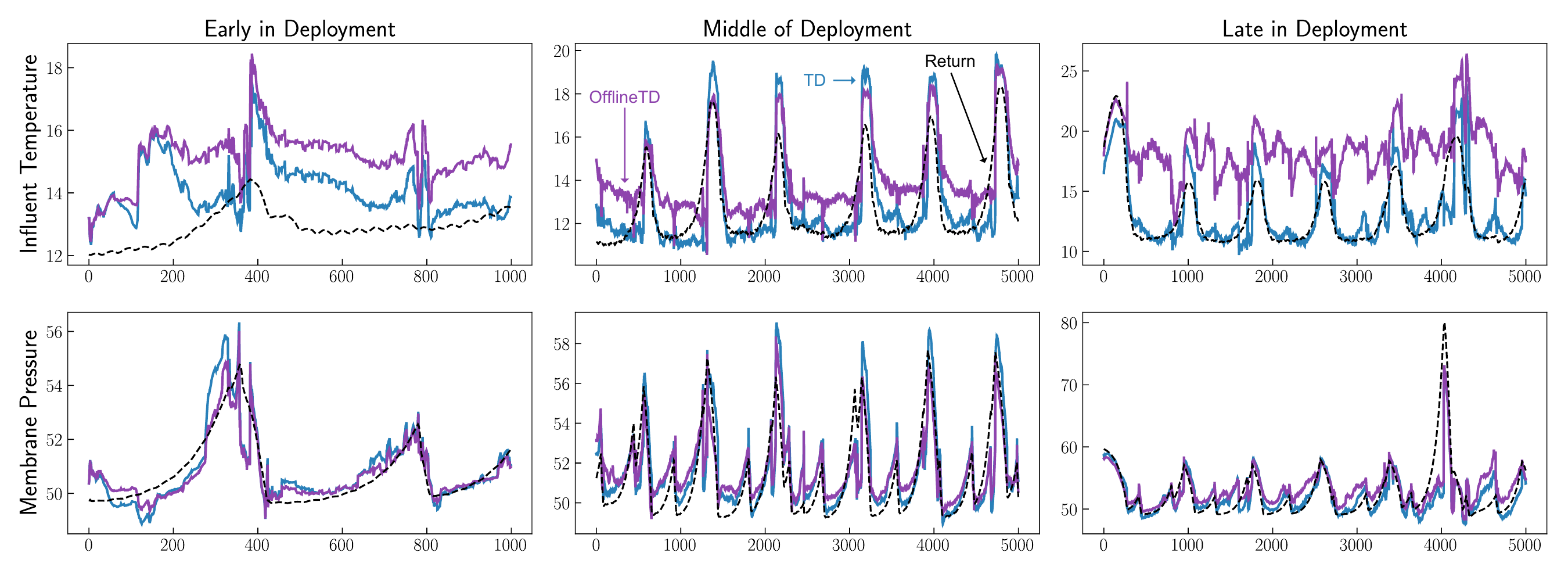}
	\caption{{\bf OfflineTD and OnlineTD predictions of Influent Temperature and Membrane Pressure on deployment data a week after training data had ended.} We sub-sample the data at the rate of 1 sample per 10 time-steps.  Both agents are trained offline on data from November 1st to November 23rd and validated using the data from November 24th to November 30th. The deployment data was taken from a 7-day period from December 1st to December 7th. Mirroring previous figures, the x-axis reports time steps (in tens of seconds) in deployment. In this case, we expect a significant distribution shift between training and deployment data. The result clearly shows this, both predictors start off far from the ideal target (Return); {\em predictions updated online in deployment (OnlineTD) can close the gap.}}
	\label{exp:offtdVStd}
\end{figure}

To highlight the need for online learning and demonstrate how changes in the data can significantly impact non-adaptive approaches, we used a dataset of 23 days for training, 7 days of validation followed by the next 7 days treated as the deployment phase.  In order to reduce the size of this dataset, we sub-sample to a rate of once for every 10 seconds, rather than every second. This also makes the prediction task harder: now a 100 step prediction corresponds to 1000 seconds. Figure \ref{exp:offtdVStd} compares GVF predictions of a frozen pre-trained agent with the online TD agent that was pre-trained on the sample data but continues to learn in deployment. Due to the differences in the training data and the deployment data, both predictors start far from the ideal prediction (the return), but only online TD can adjust as shown in the first subplot. Throughout the remainder of the deployment data online TD predictions continued to match their targets. 

This result not only highlights when online methods can be beneficial but also mimics a fairly realistic deployment scenario. Oftentimes, when working with real systems, we cannot always access the most recent data. An industrial partner might have limited data logs; or sometimes technical problems cause logs to be lost. In our specific application, water treatment, the training data might be out of date because the plant could have been out of commission. Regardless of the reason, it is useful that simple online methods like TD can adapt to such situations.

%\colorbox{yellow}{haseeb: suggestion: delete next 2 sentences?} Additionally, for the result in Figure \ref{exp:offtdVStd}, we pre-trained the predictions on 23 days of data instead of four. We made this choice to give the pre-trained agent the best chance to extract as much as possible from a wide range of operating conditions represented by nearly a month of data. 
We pre-trained the Influent Temperature predictions for 25,000 epochs and that of Membrane Pressure for 50,000 epochs. Due to resource constraints, the hyper-parameter selection is done differently in this experiment: we swept over offline learning rates $\eta\in \{1 \textsf{x} 10^{-5}, 1 \textsf{x} 10^{-6}, 1 \textsf{x} 10^{-7}\}$ and saved the network with the lowest normalized MSE on the training data. Afterwards, we used the best saved network and swept over 20 online learning rates $\alpha \in [1\textsf{x} 10^{-3}, ..., 2\textsf{x} 10^{-8}]$ range generated through a geometric progression with a common ratio of $0.5$ on the validation data.

\section{Conclusion and Future Work}
In this paper we took the first steps toward optimizing and automating water treatment on a real plant. Before we can hope to control such a complex industrial facility, we must first ensure that learning of any kind is feasible. This paper represents such a feasibility study. We provided extensive visualization and analysis of our plant's data, highlighting how it generates a large, high-dimensional data stream that exhibits interesting structure at the second, minute, day, and month timescales. Unlike the data commonly used in RL benchmarks, ours is subject to seasonal trends, and mechanical wear and tear, making it highly non-stationary. Through a combination of feature engineering and extensive offline pre-training on the operator data, we were able to learn accurate multi-step predictions encoded as GVFs. Compared with classical n-step methods used in time-series predictions, the GVF predictions were more accurate and could be learned incrementally in deployment.

The next steps for this project involve control: automating subproblems within water treatment. There are numerous such subproblems, for example, controlling the rate at which chemicals are added in pre-treatment. Backwashing is also promising because it is, by far, the most energy-intensive part of the operation. We could control the duration of backwashing or how often to backwash. At the lowest level, we can adapt the parameters of the PID controllers that control the pumps during backwashing. Classical PID controllers are not sensitive to the state of the plant; they are tuned when the plant is first commissioned and can become uncalibrated over time.

Algorithmically, we plan to investigate using our learned predictions for control, directly. Traditionally, one would define a reward function and use a reinforcement learning method such as Actor-Critic to directly control aspects of the plant operation. These methods are notoriously brittle and difficult to tune. A more practical approach is to use the predictions to directly build a controller. Prior work has explored using learned predictions inside basic if-then-else control rules to control mobile robots \cite{modayil2014prediction}. The advantage of this approach is that the control-rules are easy to explain to human operators, but since control is triggered by predictions that are continually updated in deployment the resultant controller adapts to changing conditions. An extension of this idea is to use GVF predictions---like the ones we learned in this work---as input to a neural-network based RL agent, similarly to how it was done for autonomous driving \cite{graves2020learning,jin2022offline}. 
This work provides the foundations for these next steps in industrial control with RL. 
%Given the foundations established in this paper there appears a direct, feasible path toward large-scale industrial control with RL.

% MARTHAC: This isn't happening
%\section{On Utility of Advances in Off-Policy RL}\label{sec:offpolicy}
%\begin{enumerate}
%	\item talk about different off-policy approaches
%	\item briefly discuss implementation and how they should be used
%	\item compare results and talk about why we chose a specific off-policy algorithm
%\end{enumerate}
\section{Declarations}
\paragraph{Funding}This work was supported by the Natural Sciences and Engineering Research Council of Canada (NSERC) Discovery Grant program and the Canada CIFAR AI Chairs program. Computational resources provided by Digital Research Alliance of Canada.
  
\paragraph{Conflicts of interest/Competing interests}Not applicable.

\paragraph{Ethics approval}Not applicable.

\paragraph{Consent to participate}Not applicable.

\paragraph{Consent for publication}Not applicable.

\paragraph{Availability of data and material}We are working with our industrial partner to open-source the data. This requires approval from several levels including the town municipal government. We hope to have this done by camera ready.

\paragraph{Code availability}We are working with our industrial partner to open-source the code. This requires approval from several levels including the town municipal government. We hope to have this done by camera ready. Naturally, we cannot open source any code involved in commercialization, and thus we are extracting the algorithmic code for release.

\paragraph{Authors' contributions}All authors made significant contributions to the project. {\bf Janjua}, {\bf Shah}, and {\bf Miahi} wrote all the code and ran all the experiments. {\bf White}, {\bf White} and {\bf Machado} lead the project, helped write the paper and make plots, and funded the work. All authors contributed to writing the paper.

\newpage
%%%% appendix goes here if need be
\bibliography{sn-bibliography}% common bib file

\begin{appendices}

\newpage
\section{Details on Construction of State}\label{app_state}
As discussed in Section \ref{sec:data_description}, learning directly on the raw data from a Water Treatment Plant (WTP) is very challenging due to the noisy, stochastic and partially-observable nature of the data.
In addition to this, different sensors operate at different timescales and frequencies; we summarize some of the sensors in Table \ref{tab:sp_table}. 
In order to minimize the effect of these issues on the predictions, we take a series of preprocessing steps on the raw data that are described in the sections below.

Note that we do not have significant missing data issues. 
Our system rarely misses sensor readings. However, in the rare case where we do have a missing value, we simply use zero-imputation and fill in the missing values with zeros. 

%\subsection{Handling Missing Data} 
%Our system rarely misses sensor readings. However, in the rare case where we do have a missing value, we simply use zero-imputation and fill in the missing values with zeros. 
%%In some cases, the observation vector contains missing values for some of the sensors.
%%This is situation is to be expected when working with the WTP data since the sensors can malfunction at any time.Additionally, this can also happen when the plant is in a state which is unrelated to the specific sensor.
%%We handle this problem through zero-imputation, which refers to filling in the missing values with zeros.

\begin{table}[!h]
\centering
\caption{Summary of a few sensors measuring pump speeds, setpoints valves, blowers, and PID control. All of these combine to form the input to our learning system.}
\label{tab:sp_table}
\begin{tabular}{|l|l|}
\hline
\multicolumn{1}{|c|}{\textbf{Sensor Name}} & \multicolumn{1}{c|}{\textbf{Measures}} \\ \hline
Feed Flow PID & PID control for feed flow \\ \hline
Pump Flow PID & PID control for feed/drain pump flow \\ \hline
Permeate Pump Flow PID & PID control for permeate pump flow \\ \hline
Feed Water Sample & \begin{tabular}[c]{@{}l@{}}Condition of feed water sampling valve, \\ indicating if it is open or not\end{tabular} \\ \hline
Post Flocculation Sample & \begin{tabular}[c]{@{}l@{}}Condition of post flocculation sample isolation valve, \\ indicating if it is open or not\end{tabular} \\ \hline
\begin{tabular}[c]{@{}l@{}}Process/Permeate Pump \\ Control Speed Output\end{tabular} & Speed control for process/permeate pump \\ \hline
\begin{tabular}[c]{@{}l@{}}Sulphuric Acid Pump \\ Dose Speed\end{tabular} & Speed of sulphuric acid pump dosing \\ \hline
Hypochlorite Pump & Hypochlorite pump dosing \\ \hline
\begin{tabular}[c]{@{}l@{}}Sodium Hydroxide Pump\\ Dose Speed\end{tabular} & Sodium hydroxide pump dosing \\ \hline
Citric Acid Pump & Citric acid pump dosing \\ \hline
Feed Inlet Valve & \begin{tabular}[c]{@{}l@{}}Condition of feed inlet valve, \\ indicating if it is open or not\end{tabular} \\ \hline
Feed/Waste Pump Inlet & \begin{tabular}[c]{@{}l@{}}Condition of feed/waste pump inlet valve,  \\ indicating if it is open or not\end{tabular} \\ \hline
Feed/Waste Pump Outlet & \begin{tabular}[c]{@{}l@{}}Condition of feed/waste pump outlet valve, \\ indicating if it is open or not\end{tabular} \\ \hline
\begin{tabular}[c]{@{}l@{}}Membrane Tank \\ Outlet Valve\end{tabular} & \begin{tabular}[c]{@{}l@{}}Condition of membrane tank outlet valve, \\ indicating if it is open or not\end{tabular} \\ \hline
\begin{tabular}[c]{@{}l@{}}Membrane Tank \\ Recirculation Valve\end{tabular} & \begin{tabular}[c]{@{}l@{}}Condition of membrane tank recirculation valve, \\ indicating if it is open or not\end{tabular} \\ \hline
\begin{tabular}[c]{@{}l@{}}Permeate Pump \\ Recirculation Valve\end{tabular} & \begin{tabular}[c]{@{}l@{}}Condition of permeate pump recirculation valve, \\ indicating if it is open or not\end{tabular} \\ \hline
Permeate Outlet Value & \begin{tabular}[c]{@{}l@{}}Condition of permeate outlet valve, \\ indicating if it is open or not\end{tabular} \\ \hline
BP/CIP Tank Inlet Valve & \begin{tabular}[c]{@{}l@{}}Condition of cleaning tank inlet valve, \\ indicating if it is open or not\end{tabular} \\ \hline
\begin{tabular}[c]{@{}l@{}}BP/CIP Tank \\ Recirculation Valve\end{tabular} & \begin{tabular}[c]{@{}l@{}}Condition of cleaning tank recirculation valve, \\ indicating if it is open or not\end{tabular} \\ \hline
Blower Inlet Valve & \begin{tabular}[c]{@{}l@{}}Condition of inlet blower’s valve (A/B/C), \\ indicating if it is open or not\end{tabular} \\ \hline
\begin{tabular}[c]{@{}l@{}}Membrane Aeration Blower \\ Control Speed Output\end{tabular} & \begin{tabular}[c]{@{}l@{}}Control speed output of membrane aeration blower\end{tabular} \\ \hline
Aeration Controller & Mode of the aeration (cyclic, constant, etc.) \\ \hline
Plant Mode & Mode of the plant (production, backwashing, etc.) \\ \hline
\end{tabular}
\end{table}

%In principle, water treatment plants are dependent on operational cycles to function properly, consequently many of the patterns in the telemetric data arise as the plant loops through the normal cycle. Additionally, we did establish how several external factors also affect the data stream and the patterns that arise. In real-world, it is not possible to encounter stationary, noiseless and fully observable data given the stochasticity and functioning of the world as an environment. WTPs are no different to such issues which makes constructing the agent state challenging. In addition to noise and partial observability in the data, there exist issues pertaining to how sensors operate at different frequencies and timescales. While, by design, nexting predictions can be furnished at different timescales, we take series of steps to construct the agent state to alleviate (or minimize) the effect these issues have on the raw data.
\subsection{Categorical Observations}
Some of the observations are recorded in the form of discrete categorical variables as opposed to continuous real numbers.
One example is the observation which records the current mode of the plant. 
For such observations, we encode them in a one-hot vector format.
For a categorical observation which can only take on values from one of \(k\) categories: we convert it into a binary vector of size \(k\) in which only the corresponding index of the category is set to \(1\).
%Consider a simple example of an observation $\mathcal X$ which can only take on one the values from the set $\{1,2,3\}$.
%In this case, the one-hot encoded observation will be a binary vector of size 3, where 

\subsection{Data Normalization} 
Since different sensors have different ranges, we normalize their values into the [0,1] range.
For each individual sensor value $o_{t}^{[i]}$ in the observation vector, we compute the minimum ${o}_{min}^{[i]}$ and maximum ${o}_{max}^{[i]}$ using the logs over the duration of a previous year, where $i$ is an index within the observation vector.
Afterwards, for every discrete time-step $t=1,2,\dots,$ in our dataset, we compute the normalized sensor value ${o'}_{t}^{[i]}$ as:
\begin{equation}
{o'}_{t}^{[i]} = \frac{{o}_{t}^{[i]}-{o}_{min}^{[i]}}{{o}_{max}^{[i]} - {o}_{min}^{[i]}}
\end{equation}

\subsection{Encoding Time of Day}
The observations contain the information regarding the current time of the day in seconds.
This is important since there are certain events that happen at a particular time of the day.
Additionally, there are some events that are repeated at regular intervals.
Let $\mathcal S = [s_0, s_1, s_2, \dots]$ denote the time-stamp in seconds for that day, then we encode it using sine and cosine transforms as:
\begin{align}
	s_t^{(sin)} &= \text{sin}\left(\frac{2\pi s_t}{86400}\right) \\
	s_t^{(cos)} &= \text{cos}\left(\frac{2\pi s_t}{86400}\right)
\end{align}
where 86,400 is the total number of seconds present within a day. It is the maximum value that $s_t$ can take.

\subsection{Encoding Plant Mode Length}
Understanding which mode the plant is in, and when the mode change will happen is crucial for the agent. This information is only available as a binary indicator, as mode value \(1\) against a certain mode indicates that the plant is currently in this mode, while it is \(0\) otherwise. This limitation to binary indication adds to the partial observability inherent in the state-space. We find that cyclically encoding the mode provides extra information that alleviates this associated partial observability. Since the agent has access to when the mode starts and ends as binary indicators, we utilize it to construct a cyclical thermometer encoding of the mode. 

For each mode indicator in our observation vector, we define two thermometers $w_{sine}^{[i]},  w_{cos}^{[i]} \in \mathbb R^7$ initialized to zeros, and the total mode length  $m_{l}^{[i]} \in \mathbb R$. Let \(s\) be the timestamp in seconds. Since the mode is characterized with respect to an observation that is observed at a certain time-step, we avoid explicitly denoting mode length with time-step for clarity. At each time-step, the thermometers get filled up as:

\[
w_{sine}[j] = \text{sin}\left(2^{j}\pi\left(\frac{s}{\(m_{l}^{[i]}\)}\right)\right), \\
w_{cos}[j] = \text{cos}\left(2^{j}\pi\left(\frac{s}{\(m_{l}^{[i]}\)}\right)\right)
\]
These thermometers have sine and cosine waves between the start and end of each mode, and their rotations about the period increase by a factor of \(2^{j}\) at every time-step.

 %haseeb: no need to say this twice: This equips the agent with the ability to understand when the mode is going to end, or when a mode shift is expected. Unlike binary indicators, this allows gradual increase towards the end of the mode with each new observation. 

\subsection{State Approximation and Summarizing History}
In order to make use of the historical information during predictions, we compute memory traces of the observations.
The state is constructed by appending these memory traces to the original observation vector, in addition to the mode length and time of day described in the above two sections.
For each normalized observation ${o'}_{t}^{[i]}$ at time-step $t$, we compute its memory trace ${z}_{t}^{[i]}$ using:

\begin{equation}
	{z}_{t}^{[i]} = \beta {z}_{t-1}^{[i]} + (1-\beta){o'}_{t}^{[i]}
\end{equation}

\noindent where $\beta$ is the trace decay rate hyper-parameter.
All the memory traces are initialized with zeros and are updated in an incremental manner when iterating over the dataset.

\section{Additional Results}
\begin{figure}[htb]
	\centering
	\includegraphics[width=\textwidth]{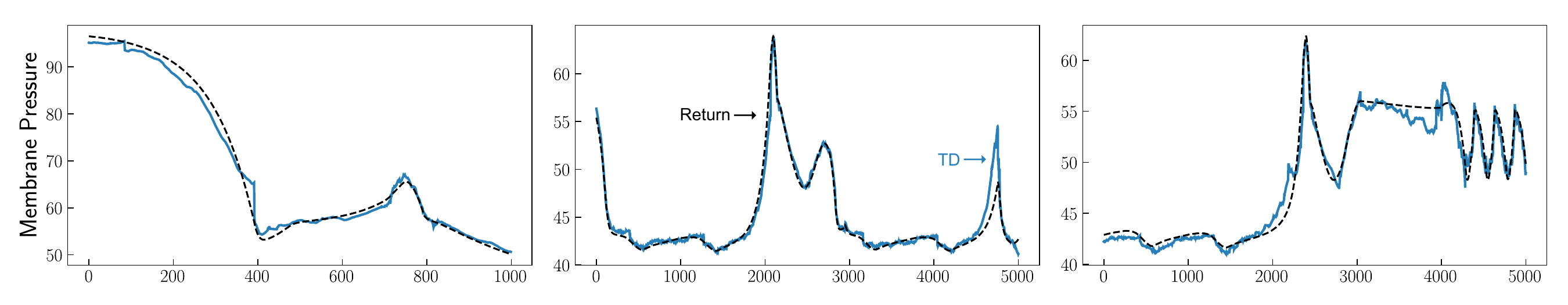}
	\caption{Predictions of the filter membrane pressure roughly 100 seconds into the future {\bf using data from May}. This plot mirrors Figure \ref{exp:result1}. The architecture, preprocessing, learning algorithms, GVF prediction, and sensor predicted are all the same. The only difference is that Figure \ref{exp:result1} reports the predictions learned from data from the month of November, and this plot uses data from May. As before, the TD predictions tightly match the target's pattern in all three scenarios.}
	\label{exp:result1_may_data}
\end{figure}

In order to verify that our results are not dependent on the time of year that we use for training, we perform an additional experiment with the same experimental setup that was used to produce Figure \ref{exp:result1}. The only difference between the experiment in Figure \ref{exp:result1} and Figure \ref{exp:result1_may_data} is that the former is trained on the data from November, 2022 whereas the latter is trained on the data from May, 2023. These results show that the predictions closely match the  return even when the data is taken from a different period of time.

\begin{figure}[htb]
	\centering
	\includegraphics[width=\textwidth]{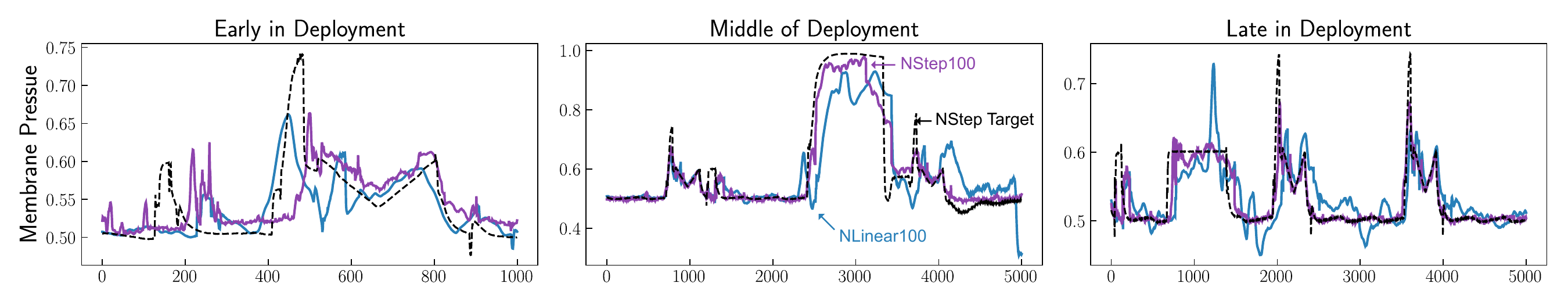}
	\caption{N-step predictions of the filter membrane pressure roughly 100 seconds into the future learned using NLinear. We show three snippets of the deployment data. The first subplot shows (on the x-axis) 1000 time-steps (seconds) at the beginning of deployment. The middle subplot shows data during a maintenance clean, and the last subplot features data near the end of the deployment phase (24 hours later). Each subplot highlights a different characteristic pattern in pressure change. The blue curve shows the n=100 step prediction learned by NLinear: first trained offline, then updated in deployment. The return represents the ideal n-step prediction and is plotted in black. The blue curve shows the n=100 step prediction learned by our non-linear n-step baseline described in Section \ref{sec:nstep}. }
	\label{exp:NLinear}
\end{figure}

In Figure \ref{exp:NLinear}, we evaluate NLinear and compare it with the N-step method the exact same data as in Figure \ref{exp:result1}. We used a prediction horizon of n=100 and a history length (look-back window) of 4000 time-steps. We compare the NLinear results with the N-step since both of these methods share the same prediction target i.e. the cumulant value that is 100 steps into the future. These results showcase the limitations of the NLinear methods: they can only make correct predictions when there is a clear periodicity within the data. Additionally, the look-back window size needs to be large enough so that the linear layer can make future predictions according to the repeated past patterns. As a result of this, linear methods are simply insufficient for learning a good prediction model on our current dataset.

\section{Comparing to TD with Replay}\label{app_replay}

\begin{figure}[htb]
	\centering
	\includegraphics[scale=0.2]{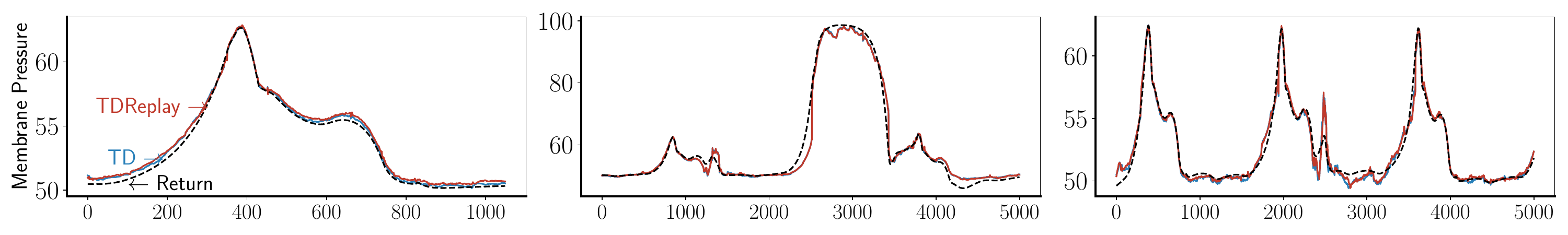}
	\includegraphics[scale=0.2]{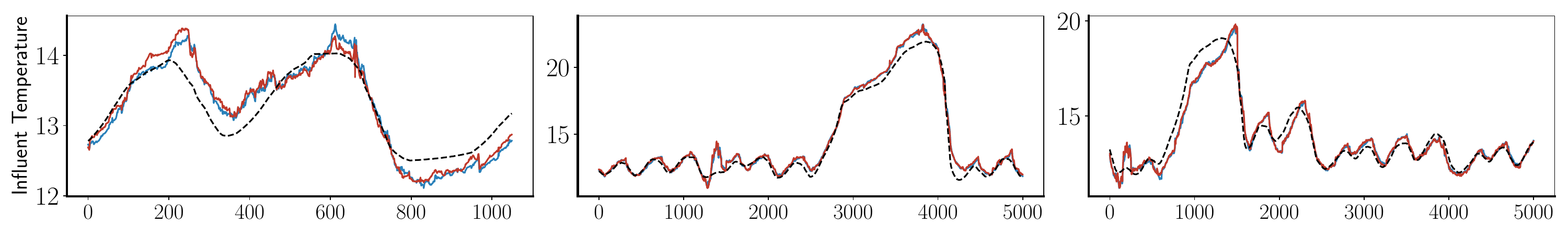}
	\includegraphics[scale=0.2]{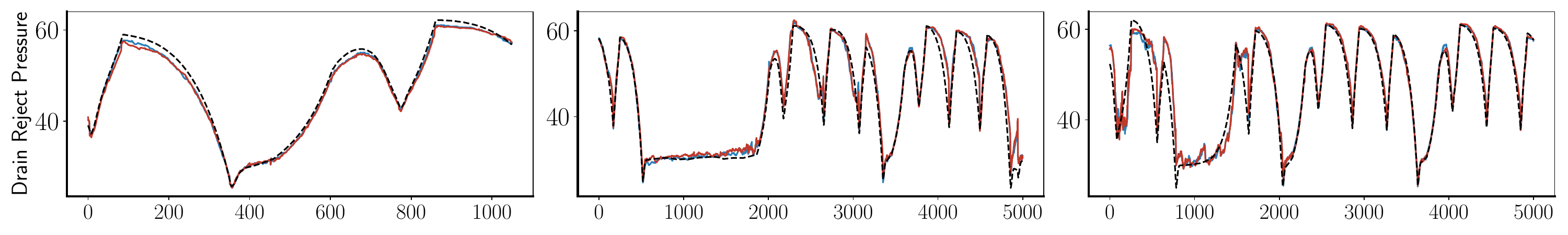}
	\includegraphics[scale=0.2]{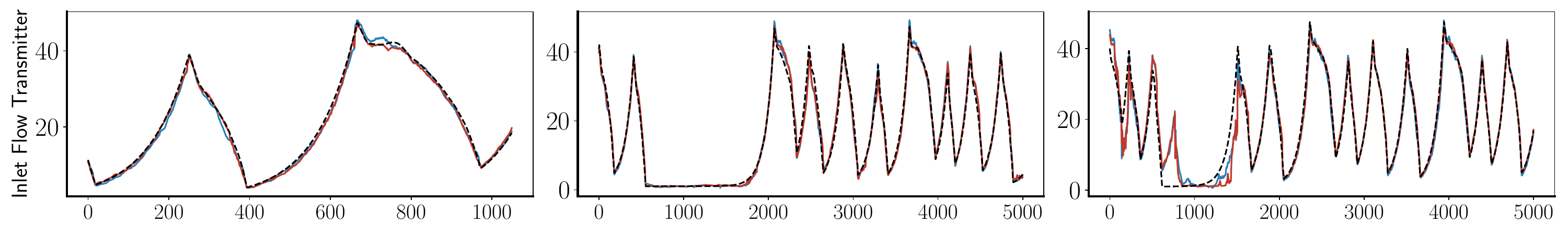}
	\includegraphics[scale=0.2]{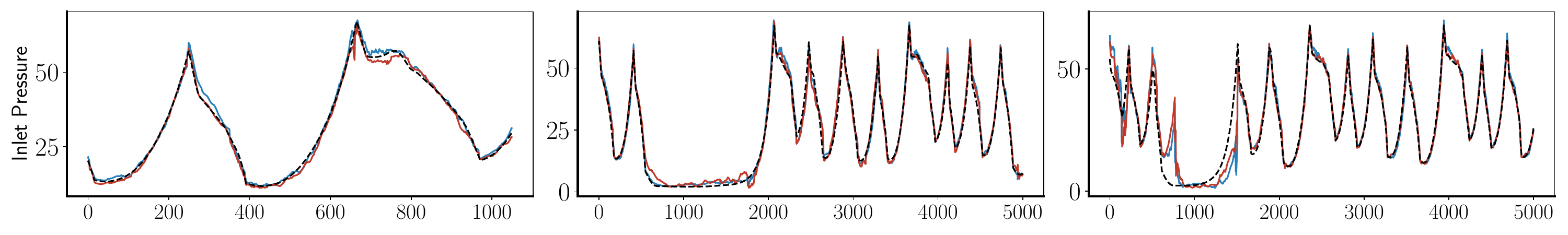}
	\caption{Comparing online TD and online TD with replay for GVF Predictions across several sensors. The structure of this plot mirrors Figure \ref{exp:nstep}.   }
	\label{exp:allalgs}
\end{figure}

We considered both the simpler online TD update in deployment, as well as using TD with replay. The TD with replay algorithm is summarized in Algorithm \ref{alg:tdro}. We found, though, that they performed very similarly (see Figure \ref{exp:allalgs}), so we used the simpler online TD update in the main body.

\begin{algorithm}[!htb]
\caption{TDwithReplay using Offline Pretraining}\label{alg:tdro}
\begin{algorithmic}[1]
	\State Hyperparameters: offline stepsize $\eta > 0$, batchsize $k$, number of epochs $n_{\text{epochs}}$, online stepsize $\alpha > 0$, number of replay steps $n_{\text{replay}}$
	\State Input \(\mathcal{D}_{\text{augmented}} = \{(\shat_{t}, c_{t+1}, \shat_{t+1})\}\) 
	\State $\vect{w_{0}}, s_{\text{opt}} = $ OfflineTD($\mathcal{D}_{\text{augmented}}$, $\eta$, $k$, $n_{\text{epochs}}$)
	 \State Initialize the replay buffer \(\mathcal{B}\) with last $kn_{\text{replay}}$ samples in $\mathcal{D}_{\text{augmented}}$
        \State Obtain initial observation $o_t$ for $t = 0$, set $\shat_0 = o_0$
        \While{\text{in deployment}}
        \State Observe next observation $o_{t+1}$ and cumulant $c_{t+1}$
         \State \(\shat_{t+1} \leftarrow U(o_{t+1}, \shat_t)\) 
            \State \(v_{t+1} \leftarrow f_{\params_{t}}(\shat_{t+1})\) 
            \State \(\delta_{t} \leftarrow c_{t+1} + \gamma v_{t+1} - f_{\params_{t}}(\shat_{t})\) 
            \State \(\params \gets \params_{t} + \alpha \delta_{t}\nabla f_{\params_t}(\shat_{t})\) 
             \State Add tuple $(\shat_{t}, c_{t+1}, \shat_{t+1})$ to $\mathcal{B}$
             \For{$n_{\text{replay}}$ steps}
             \State Sample a random mini-batch \verb+batch+ of size $k$ from $\mathcal{B}$
            \State $\Delta \leftarrow -\frac{1}{k}\sum_{i \in \text{batch}} \left( c_{i+1} + \gamma f_{\vect{w_t}}(\shat_{i+1}) - f_{\params}(\shat_{i}) \right)\nabla f_{\params}(\shat_{i})$
            \State $\params, s_{\text{opt}} \gets \verb+opt+(\params , \Delta, \alpha, s_{\text{opt}})$
              \EndFor
              \State $\params_{t+1} = \params$
            \State $t = t+1$
 	   % \State \(\shat_{t} \leftarrow \shat_{t+1}\)
        \EndWhile
    \end{algorithmic}
\end{algorithm}

\end{appendices}

%%===========================================================================================%%
%% If you are submitting to one of the Nature Portfolio journals, using the eJP submission   %%
%% system, please include the references within the manuscript file itself. You may do this  %%
%% by copying the reference list from your .bbl file, paste it into the main manuscript .tex %%
%% file, and delete the associated \verb+\bibliography+ commands.                            %%
%%===========================================================================================%%

%% if required, the content of .bbl file can be included here once bbl is generated
%%\input sn-article.bbl

%% Default %%
%%\input sn-sample-bib.tex%

\end{document}